\documentclass[10pt,twocolumn,letterpaper]{article}

\usepackage{cvpr}              % To produce the CAMERA-READY version
\usepackage{graphicx}
\usepackage{amsmath}
\usepackage{amssymb}
\usepackage{booktabs}
\usepackage{multirow}
\usepackage{soul}
\usepackage{changepage,threeparttable}
\usepackage[accsupp]{axessibility}
\usepackage{mmstyle}
\usepackage[table,xcdraw,dvipsnames]{xcolor}
\usepackage[most]{tcolorbox}
\usepackage{listings}
\usepackage{lipsum}
\usepackage{titletoc}
\usepackage{longtable}
\newcommand{\tablestyle}[2]{%
  \setlength{\tabcolsep}{#1}%
  \renewcommand{\arraystretch}{#2}%
  \centering
  \small
}
\definecolor{cvprblue}{rgb}{0.21,0.49,0.74}
\definecolor{iccvblue}{rgb}{0.21,0.49,0.74}
\usepackage[pagebackref,breaklinks,colorlinks,allcolors=cvprblue]{hyperref}
\definecolor{cvprblue}{rgb}{0.21,0.49,0.74}
\usepackage[pagebackref,breaklinks,colorlinks,allcolors=cvprblue]{hyperref}

\def\paperID{13034} % *** Enter the Paper ID here
\def\confName{CVPR}
\def\confYear{2026}

\title{Multi-SpatialMLLM: Multi-Frame Spatial Understanding with Multi-Modal Large Language Models}

\author{Runsen Xu$^{1,2}$\quad
Weiyao Wang$^{1}$\quad
Hao Tang$^{1}$\quad
Xingyu Chen$^{1}$\quad
Xiaodong Wang$^{1}$\\
Fu-Jen Chu$^{1}$\quad
Matt Feiszli$^{1}$\quad
Kevin J. Liang$^{1}$\quad
\\
$^1$FAIR, Meta\quad
$^2$The Chinese University of Hong Kong
}

\begin{document}
\twocolumn[{%
\renewcommand\twocolumn[1][]{#1}%
\maketitle
\begin{center}
    \centering
    \vspace{-4ex}
    \includegraphics[width=1.0\linewidth]{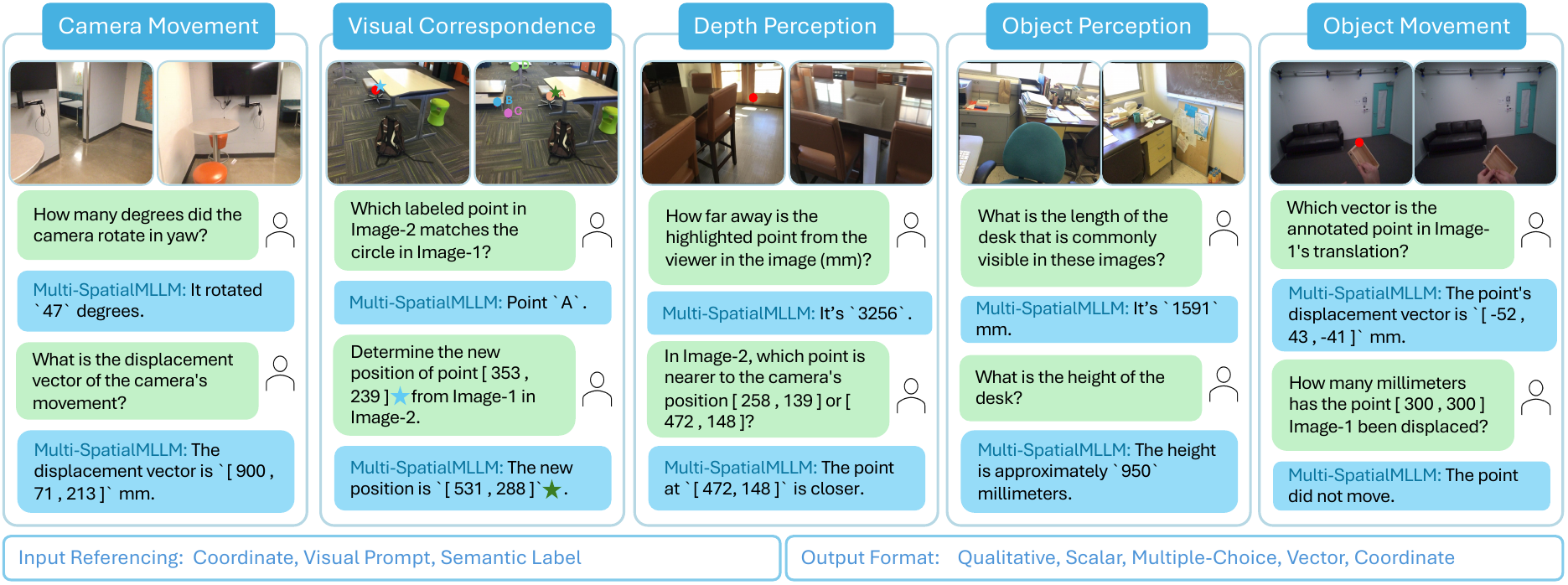}
    \captionof{figure}{We present \textbf{Multi-SpatialMLLM}, a model capable of multi-frame spatial understanding, a capability overlooked by previous spatial understanding research. Multi-SpatialMLLM can support different types of input referencing and outputs for various tasks.}
    \label{fig:teaser}
\end{center}
}]

\begin{abstract} 
Multi-modal large language models (MLLMs) have rapidly advanced in visual tasks, yet their spatial understanding remains limited to single images, leaving them ill-suited for physical-world applications that require multi-frame reasoning. In this paper, we propose a framework to equip MLLMs with multi-frame spatial understanding by integrating fundamental spatial skills, including depth perception, visual correspondence, and dynamic perception. We design a novel data pipeline and collect the MultiSPA dataset of more than 27 million samples spanning diverse 3D and 4D scenes to enable training. Alongside MultiSPA, we introduce a comprehensive benchmark that tests a wide spectrum of spatial tasks under uniform metrics. Our resulting model, Multi-SpatialMLLM, achieves significant gains over baselines and proprietary systems, demonstrating scalable and generalizable multi-frame perception. We further observe multi-task benefits and emergent spatial capabilities in challenging scenarios, and showcase how our model can serve as a multi-frame reward annotator for robotics.
\end{abstract}    
\section{Introduction}
\label{sec:introduction}
Recent years have witnessed tremendous advances in multi-modal large language models (MLLMs), which have evolved into versatile AI assistants capable of a wide array of tasks~\cite{Kosmos-1, LLaVA_OneVision, InternLM, GPT-4o, InstructBLIP}. Despite these strides, deploying such assistants as ``brains in a jar” within digital platforms limits their potential. Instead, there is a growing push to integrate MLLMs directly into real-world applications, such as robotics~\cite{RoboVLMs, Pi0} and autonomous vehicles~\cite{DriveLM}, to facilitate interactions with the environment. This shift imposes a requirement for human-like spatial understanding. However, current MLLMs struggle with surprisingly basic spatial understanding, even confusing left and right~\cite{SpaitalMM}.

\begin{table*}[t!]
\centering
\caption{\textbf{Comparison of spatial understanding datasets.} Our MultiSPA is the first large-scale dataset for multi-frame spatial understanding, with diverse referencing and output formats. We generate 27M samples here and can scale further if needed.}
\label{tab:compare_datasets}
\vspace{-0.7em}
\scalebox{0.978}{\tablestyle{3.8pt}{0.87}

\begin{tabular}{@{}lccccccc@{}}
\toprule
Dataset & Split & Multi-Frames & GT Annotation & Referencing & Output & \# Images & \# QAs \\ \midrule
BLINK~\cite{BLINK} & eval & \cmark & \cmark & dot & MCQ & 877 & 572 \\
UniQA-3D~\cite{towards3Dvision} & eval & \cmark & \cmark & dot, semantic & MCQ & 2450 & 2450 \\
Q-Spatial~\cite{Q-Spatial} & eval & \xmark & \cmark & semantic & scalar & 271 & 271 \\
VSR~\cite{VSR} & train, eval & \xmark & \cmark & semantic & true/false & 11K & 11K \\
SpatialVLM~\cite{SpatialVLM} & train, eval & \xmark & Only eval & semantic & qual., MCQ, scalar & 10M & 2B \\
SpatialRGPT~\cite{SpatialRGPT} & train, eval & \xmark & Only eval & mask & qual., scalar & 1M & 8.7M \\ \midrule
MultiSPA & train, eval & \cmark & \cmark & dot, coord., semantic & qual., MCQ, scalar, coord., vec. & 1.1M & 27M+ \\ 
\bottomrule
\end{tabular}

}
\end{table*}

Previous works~\cite{SpatialVLM, SpatialRGPT} attribute these deficiencies primarily to a shortage of specialized training data, and addresses it by incorporating spatial data into model training, leading to notable improvements. However, these works focus on single-image scenarios, thus restricting the model’s perception to a static field-of-view without any dynamic information.
We instead aim for more comprehensive spatial understanding, enabling MLLMs to reason across multiple images. Inspired by the long-standing Structure-from-Motion problem~\cite{MVG} from 3D computer vision, we focus on integrating three fundamental capabilities into MLLMs: (1) depth perception, to infer relative distances and three-dimensional structures (2) visual correspondence, to match overlapping regions across images for consistent scene association, and (3) dynamic perception, to perceive self-movement (camera motion) and object motion.

The challenge in achieving this goal is the scarcity of suitable training data. Because manual annotation at the required scale can be expensive, prior works~\cite{SpatialVLM, SpatialRGPT} have resorted to single-view data from in-the-wild images~\cite{OpenImages}, relying on off-the-shelf modules such as monocular depth estimators~\cite{Metric3Dv2} and open-vocabulary object detectors~\cite{GDINO-1.5}. However, this approach often produces noisy annotations. Moreover, our objective is to collect multi-frame spatial data, which requires both spatial and temporal alignment—an open challenge~\cite{CUT3R, yang2025fast3r} in unstructured in-the-wild images. Consequently, we leverage existing annotated 3D~\cite{ScanNet} and 4D~\cite{ADT, PStudio, TAPVid3D} datasets for data collection.

We develop a data engine that samples image pairs with a uniform overlap distribution, then backprojects spatially and temporally aligned point clouds to establish pixel correspondences. Leveraging these correspondences in addition to camera movement and projection information, we create high quality question–answer pairs via diverse, LLM-generated templates. In contrast to previous methods that rely on semantic labels~\cite{SpatialVLM} or object masks~\cite{SpatialRGPT} for referencing, our framework supports multiple modalities, including visual point annotations, pixel coordinates, and semantic labels, thus broadening potential downstream applications. The collected data encompasses both qualitative and quantitative spatial information, ranging from text to scalar, 2D pixel locations, and 3D displacement vectors.

In total, we curate a dataset named \textbf{MultiSPA} consisting of more than 27 million samples, which we use to train our \textbf{Multi-SpatialMLLM}, as illustrated in \cref{fig:teaser}. To the best of our knowledge, MultiSPA is the first large-scale dataset dedicated to multi-frame spatial understanding. Alongside the dataset, we introduce a novel \textbf{MultiSPA benchmark} to evaluate multi-frame spatial reasoning in MLLMs, covering diverse tasks and output formats under a unified metric.

Extensive experiments show that Multi-SpatialMLLM substantially outperforms base models and proprietary systems, and even matches specialized 3D perception models. We show that its multi-frame spatial understanding generalizes to held-out benchmarks and scales with data volume, task diversity, and trainable parameters, without degrading the base models’ general abilities. To further provide insights into spatial training, we find that multi-task training on MultiSPA provides notable benefits and, for the first time, reveals preliminary signs of emergent behavior on spatial tasks. Finally, we demonstrate our model’s potential as a multi-frame reward annotator for robot learning.
\vspace{-1mm}
\section{Related Work}
\label{sec:related_work}
\vspace{-1mm}
\noindent\textbf{Multi-modal large language models.}
We refer to multi-modal large language models (MLLMs) or vision-language models (VLMs) as large language models extended to handle image inputs~\cite{LLaVA1_2, MMGPT, MiniGPT-4, InternVL2_0, QwenVL, DeepSeekVL, ImageBind-LLM, VILA, InstructBLIP}, typically by incorporating an image encoder~\cite{CLIP, OpenCLIP, InternViT} converting images into tokens, which are then projected into the LLM's~\cite{LLaMA, InternLM, Qwen} text latent space and processed alongside text tokens. 
Training MLLMs uses the same language modeling objective as LLMs. Thanks to Internet-scale image–text training data, \eg captioning or OCR corpora~\cite{Laion-5b-eng, Laion-coco}, MLLMs have demonstrated remarkable performance on tasks like multi-modal dialogue. However, these training data lack sufficient spatial annotations, resulting in deficient spatial understanding. We address this limitation by further fine-tuning existing MLLMs on newly collected spatial data. Notably, we preserve the original model architecture to maintain the wide range of capabilities and application scenarios derived from large-scale pre-training.

\noindent\textbf{Spatial understanding benchmarks for MLLMs.}
Researchers have explored deploying MLLMs on real-world platforms such as robotics~\cite{PaLM-E, OpenVLA, Pi0, RoboVLMs,VLM-Grounder} and autonomous vehicles~\cite{DriveLM}, applications requiring human-like spatial understanding to perceive and interact with the environment. However, spatial understanding is a complex, fundamental ability that is difficult to define formally, so researchers have introduced various benchmarks targeting different aspects of spatial reasoning for evaluation purposes. Most prior works focus on single-image spatial understanding, primarily assessing inter-object spatial relations~\cite{Q-Spatial, SpaitalMM, SPHERE, CVBench, VSR, CLEVR, Embspatial} or spatial recognition inspired by cognitive science~\cite{SpaceRecognition}. Some benchmarks extend beyond single images: BLINK~\cite{BLINK} and UniQA-3D~\cite{towards3Dvision} evaluate spatial relationships across image pairs, while video-based benchmarks~\cite{OST-Bench,SeeAcrossViews} like VSI-Bench~\cite{VSI-Bench} introduce scene-level spatial reasoning. Though these share some similar tasks with ours, such as qualitative camera movement estimation and keypoint matching, our proposed benchmark includes additional tasks like object movement perception, and supports more diverse input and output formats, instead of just the limited multiple-choice format.

\noindent\textbf{Improving MLLMs for spatial understanding.}
Existing benchmarks highlight the limitations of MLLMs in spatial understanding, prompting several recent works~\cite{SpatialVLM, SpatialRGPT, PerceptionToken, SpatialBot, RoboSpatial, SpatialPIN, SpatialMLLM, VLM3R}. Most of these focus on single-image understanding. SpatialVLM~\cite{SpatialVLM} was the first to identify the lack of spatial training data as a key limitation, demonstrating significant improvements by fine-tuning MLLMs on a curated spatial dataset. SpatialRGPT~\cite{SpatialRGPT} extended this approach by introducing mask-based reference and incorporating depth images. SpatialPIN~\cite{SpatialPIN} explored an alternative strategy, avoiding model fine-tuning and instead leveraging specialized perception models to extract spatial information for MLLMs. Unlike prior efforts, we focus on enabling MLLMs to reason across multiple images for spatial understanding by fine-tuning them on our newly collected dataset. Concurrently, SAT~\cite{SAT} also explores multi-frame spatial reasoning, but it relies on simulated data, potentially introducing a sim-to-real gap. In contrast, our dataset is significantly larger, derived from real-world images across diverse scenarios, and covers a broader range of spatial reasoning tasks. See \cref{tab:compare_datasets} for a comparison of our dataset with other popular spatial datasets and benchmarks.
\section{MultiSPA Dataset and Benchmark}
\label{sec:method}

In this section, we introduce our MultiSPA dataset in \cref{sec:multispa_overview}, describe the data generation pipeline in \cref{sec:multispa_dataset}, and present the MultiSPA benchmark in \cref{sec:multispa_benchmark}.

\subsection{MultiSPA Dataset}
\label{sec:multispa_overview}
\noindent\textbf{Tasks definitions.}
We aim to equip MLLMs with the ability to integrate multiple images for spatial understanding. Building on the three fundamental capabilies discussed in \cref{sec:introduction}, we introduce the following five tasks to generate training data: 1) Depth perception, 2) Visual correspondence, 3) Camera movement perception, 4) Object movement perception, and 5) Object size perception. \cref{fig:teaser} shows examples of these five tasks.

\noindent\textbf{Referencing and output types.}
As summarized in \cref{tab:compare_datasets}, we reference specific pixels or objects in spatial QA data using visual dots (points) or semantic labels, as opposed to masks\cite{SpatialRGPT} to avoid additional dependency on a segmentation module~\cite{SAM}. Additionally, we introduce pixel coordinates as a straightforward referencing method that preserves the original images without requiring extra annotations. Beyond referencing, most existing datasets constrain spatial tasks to multiple-choice formats or limit outputs to qualitative or scalar quantitative answers. We broaden these restrictions by incorporating diverse quantitative outputs such as pixel coordinates and displacement vectors. Detailed descriptions of each task are provided as follows, with examples of each in the supplementary material.

\noindent\textbf{Depth perception.}
We divide this task into two subtasks: direct depth estimation and depth comparison. In the first task, a single pixel is specified in an image, and the model must estimate its depth. In the second, two pixels are specified, and the model must identify the one closer to the camera. Both subtasks support referencing pixels either via visually annotated dots or pixel coordinates.

\noindent\textbf{Visual correspondence.}
Given two images and a pixel location in the first image, the model must identify the corresponding pixel with the same 3D position in the second image, either qualitatively or quantitatively. In the qualitative version, the pixel is annotated visually in both images, and an additional three pixels in the second image are labeled to form a multiple-choice question. The model's goal is to pick the correct label. In the quantitative version, only the pixel coordinates are specified, and the model must output the corresponding coordinates in the second image.

\noindent\textbf{Camera movement perception.}
Given two images, the model must estimate the relative movement of the camera from the first view to the second, including both translation and orientation. We define multiple output levels, from coarse to fine-grained. In the simplest variant, the model must only identify the camera's movement direction along three translational axes: ``right" or ``left", ``forward" or ``backward", and ``up" or ``down", as well as its rotation direction in two axes: rotating ``left" or ``right" and tilting ``up" or ``down". A more challenging variant requires the model to estimate scalar values of the overall translation distance or rotation angle. Finally, the most detailed form requires predicting the camera's displacement vector. In total, we have nine question types for this category.

\noindent\textbf{Object movement perception.}
Given two images and a pixel location on a specific object (or object part) in the first image, the model estimates the pixel's overall translation distance or, at a finer level, the pixel's displacement vector with respect to the first view. The camera may remain still or move during capture, and pixel referencing can be done either via visual annotations or pixel coordinates.

\noindent\textbf{Object size perception.}
Given several images of a target object, the model estimates the object’s height, width, and length. We treat this as a higher level of spatial understanding compared with the previous four tasks. The model must integrate information across all images to infer the object's size. We use semantic labels to refer to the object.

\subsection{MultiSPA Data Generation}
\label{sec:multispa_dataset}

\noindent\textbf{Data Format.}
Following common MLLM fine-tuning strategies~\cite{LLaVA1_5,InternVL2_0,InstructBLIP}, we format our data as QA pairs as:

\begin{quote}
\textbf{User:} \textcolor{bluecolor}{\textless image\textgreater...\textless image\textgreater}\textcolor{customred}{\{description\}\{question\}}

\textbf{Assistant:} \textcolor{customgreen}{\{answer\}}
\end{quote}
We use GPT-4o~\cite{GPT-4o} to generate diverse templates for task descriptions, questions, and answers. Please refer to the supplementary material for detailed templates for each task. To facilitate answer extraction, we enclose the answer in backticks (\texttt{``}). For numerical answers of metric length, we use millimeters as the unit and round to the nearest integer. For pixel coordinates, we normalize the values to maintain compatibility with varying image resolutions as follows:
\begin{equation}
    x_{\text{norm}} = \left\lfloor \frac{x}{W} \times 1000 \right\rfloor,
    y_{\text{norm}} = \left\lfloor \frac{y}{H} \times 1000 \right\rfloor
\end{equation}
where \(x, y\) are the original pixel coordinates, and \(W, H\) are the width and height of the image, respectively.

\noindent\textbf{Source datasets.}
We leverage existing annotated scene datasets for high-quality data collection. Specifically, we use the 4D datasets Aria Digital Twin (ADT)~\cite{ADT} and Panoptic Studio (PStudio)~\cite{PStudio}, with 3D tracking annotations from the TAPVid3D~\cite{TAPVid3D} dataset for the object movement perception task, and the 3D dataset ScanNet~\cite{ScanNet} for other tasks. Our data generation pipeline can be used for other datasets as long as they have the same spatial annotations. Further details are in the supplementary material.

\subsubsection{Static Scene Data Generation}
\noindent\textbf{Visible points calculation.}
For each scene, ScanNet~\cite{ScanNet} provides a reconstructed point cloud 
$\mathcal{P}_{\text{scene}}=\{\mathbf{p}^W\}$, where each point $\mathbf{p}^W=(X,Y,Z)^T$ is in world coordinates. 
Each RGB image $\mathbf{I}_i$ has a depth map $\mathbf{D}_i$, an extrinsic matrix $\mathbf{E}_i$ (camera to world), and an intrinsic matrix $\mathbf{K}_i$. 
We transform and project each point $\mathbf{p}^W$ onto $\mathbf{I}_i$ via:

\begin{equation}
\mathbf{p}^C_i = (\mathbf{E}_i)^{-1}
\begin{bmatrix}
\mathbf{p}^W \\ 1
\end{bmatrix},
\quad
\begin{bmatrix}
u \\ v \\ 1
\end{bmatrix}
=
\frac{\mathbf{K}_i}{\mathbf{p}^C_i[2]}
\begin{bmatrix}
\mathbf{p}^C_i[0] \\ 
\mathbf{p}^C_i[1] \\ 
\mathbf{p}^C_i[2]
\end{bmatrix}.
\label{eq:proj}
\end{equation}
We maintain all visible points of image $i$, denoted as $\mathcal{P}_i$, by selecting those whose projected coordinates $(u,v)$ lie within the image bounds and are not occluded: 

\begin{equation}
0 < \mathbf{p}^C_i[2] < \mathbf{D}_i(u,v).
\label{eq:visibility}
\end{equation}

\begin{figure}[t!]
\centering
\includegraphics[width=0.98\linewidth]{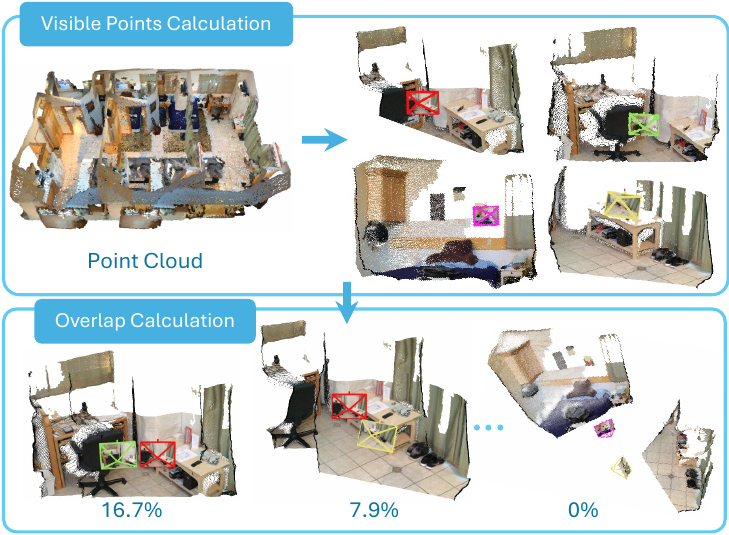}
\caption{\textbf{Overlap ratio calculation of image pairs.}}
\vspace{-4mm}
\label{fig:scannet_method}
\end{figure}

\noindent\textbf{Depth perception data generation.}
To create depth perception data, we randomly sample images for each scene. For each image $\mathbf{I}_i$, we sample one or two visible points from $\mathcal{P}_i$, record their 2D coordinates $(u,v)$ and corresponding depth $\mathbf{p}^C_i[2]$, and fill in the templates to construct QA pairs.

\noindent\textbf{Image pairs sampling.}
Although depth estimation is performed on single images, we also require image pairs with overlapping regions to construct multi-frame spatial understanding data. For each scene, we define the overlap ratio between two images as the IoU of their visible points:
\begin{equation}
    \text{Overlap}(i,j)=\frac{|\mathcal{P}_i\cap\mathcal{P}_j|}{|\mathcal{P}_i\cup\mathcal{P}_j|}.
\end{equation}
We only consider image pairs with overlap ratios between 6\% and 35\%, as ratios outside this range indicate either too little or too much shared content. \cref{fig:scannet_method} visualizes the calculation of the overlap ratio. Notably, the overlap ratio exhibits a long-tailed distribution, where most pairs have a low overlap. We do not use all image pairs, and to achieve balanced sampling, we divide the overlapping pairs into bins based on their overlap ratios. We then evenly allocate a target number of samples among these bins while prioritizing bins with fewer samples. For each task, we sample image pairs with different random seeds to ensure diversities. 
More details are in the supplementary material.

\noindent\textbf{Visual correspondence data generation.}
For an sampled image pair $(\mathbf{I}_i, \mathbf{I}_j)$, we randomly select one point from their co-visible points $\mathcal{P}_i\cap\mathcal{P}_j$ and use its projected pixel coordinates in both images to construct the QA pair.

\noindent\textbf{Camera movement perception data generation.}
In the ScanNet\cite{ScanNet} dataset, the camera coordinate system is defined with its origin at the top-left of the image, where the $x$-axis points to the right, the $y$-axis points downward, and the $z$-axis points forward. For an image pair $(\mathbf{I}_i, \mathbf{I}_j)$, we compute the relative camera pose with respect to the first image as 
$\mathbf{E}_j^i=\mathbf{E}_i^{-1}\mathbf{E}_j\in\mathbb{R}^{4\times4}.$
The translation component is given by the displacement vector 
$(\mathbf{d}_j^{i})^T=[x_j^i,y_j^i,z_j^i]=\mathbf{E}_j^i[0\!:\!3,\!3]$,
and its norm, $\|\mathbf{d}_j^i\|$, represents the overall translation distance.
When $x_j^i>0$, $y_j^i>0$, and $z_j^i>0$, we label the camera motion as ``right", ``down", and ``forward", respectively; otherwise, the movement is considered “left”, “up”, or “backward.” To determine orientation, we measure rotation angles around the gravity direction and the tilt relative to the ground plane (details in the supplementary material). Finally, we format all these spatial parameters into QA templates to construct the camera movement data.

\noindent\textbf{Object size perception data generation.}
For this task, we require a set of images that not only share overlapping regions but also jointly cover the entire target object. To ensure that the model learns to reason across all images, only the complete image set should cover the object’s full dimensions, while no proper subset does. To achieve this, we propose a BFS-based \textbf{minimum-coverage-set search} algorithm that iteratively explores image combinations with early pruning. For each object in ScanNet~\cite{ScanNet}, we use the size of the target object's 3D bounding box as its “height,” “width,” and “length” and combine these with the searched image sets to construct QA pairs. More details are in the supplementary material.

\subsubsection{Dynamic Data Generation}
TAPVid3D~\cite{TAPVid3D} provides temporally aligned point cloud tracking sequences $\{\mathcal{P}_t\}_{t=1}^{T}$, along with the corresponding video frames $\{\mathbf{I}_t\}_{t=1}^{T}$, camera extrinsics $\{\mathbf{E}_t\}_{t=1}^{T}$, and intrinsics $\{\mathbf{K}_t\}_{t=1}^{T}$ for the ADT\cite{ADT} and PStudio datasets\cite{PStudio}.
We use these datasets to construct object movement perception QA pairs. We randomly select one point from the tracked sequences, then choose two images \((\mathbf{I}_i, \mathbf{I}_j)\) to form the image pair. Similar to the camera movement data generation procedure, we compute each point's displacement vector and translation distance between these two frames using the camera extrinsics. To ensure diversity, we adopt two additional modules described as follows to sample the points and image pairs (more in supplementary material).

\begin{figure}[t!]
\centering
\includegraphics[width=0.98\linewidth]{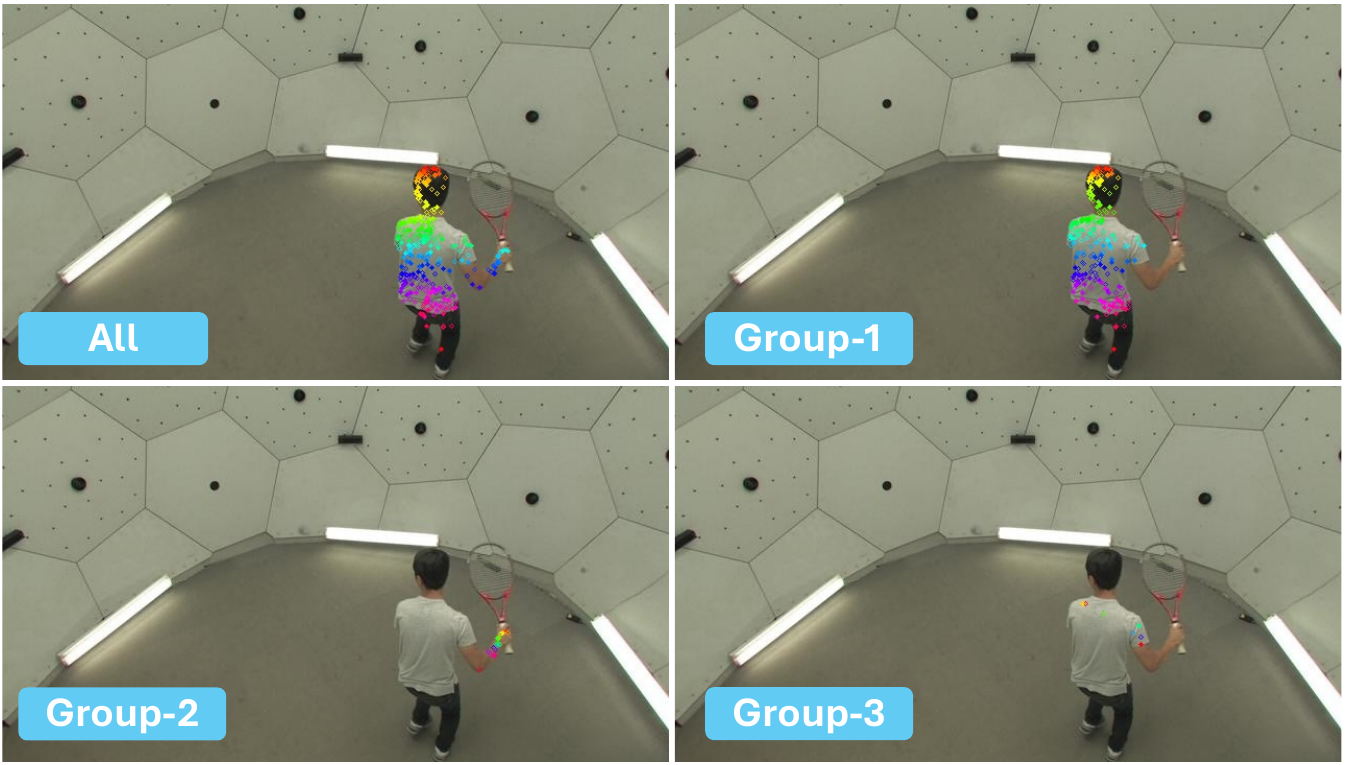}
\vspace{-1mm}
\caption{\textbf{Visualization of rigid body segmentation results.}}
\vspace{-3mm}
\label{fig:rigid_body}
\end{figure}

\noindent\textbf{Rigid body segmentation.}
Point clouds from TAPVid3D typically belong to the same object or a local region, but different parts may be unevenly represented (see \cref{fig:rigid_body}). For instance, a moving human often has more points on the torso than on the arms. Random sampling yields a distribution skewed toward the dominant body part, which follows a single movement pattern. Thus, we devise a clustering-based rigid body segmentation method to group the point clouds according to inter-point distance changes over time, and sample each group separately to enhance diversity.

\noindent\textbf{Image pairs bin sampling.}
Given a selected point that appears in \(T\) frames, one could form up to \(\frac{T(T-1)}2\) image pairs. However, similar to ScanNet, these pairs exhibit a long-tailed distribution of motion magnitudes. We therefore bin the image pairs by the object translation distances and perform balanced sampling for each bin, ensuring diversity across small and large displacements.

\subsection{MultiSPA Benchmark}
\label{sec:multispa_benchmark}
Using both ScanNet~\cite{ScanNet} and TAPVid3D~\cite{TAPVid3D}, we employ our proposed data generation pipeline to create over 27M QA samples from 1.1M unique images. For each subtask in MultiSPA, we hold out 300 samples as evaluation sets, resulting in a total of 7,800 benchmark samples. We ensure that the images in the benchmark come from scenes or scenarios distinct from those in the training split. 
The dataset distribution is provided in the supplementary material.

\noindent\textbf{Evaluation metric.}\quad
Our MultiSPA benchmark supports diverse answer formats. The required answer format is specified in the question, and a regular expression is used to extract the answer from model responses. Accuracy is calculated using task-specific criteria. For qualitative and multiple-choice answers, exact matching is used. For scalar and vector, a prediction $\mathbf{v}_{pred}$ is correct if: $\|\mathbf{v}_{pred} - \mathbf{v}_{gt}\|_2 \le 0.2 \cdot \|\mathbf{v}_{gt}\|_2$. For pixel coordinates, a prediction is correct if within 5\% image width pixels of the ground truth.
\section{Experimental Results}
\label{sec:exps}

\begin{table*}[th!]
\centering
\caption{\textbf{Evaluation on the MultiSPA benchmark.} Our Multi-SpatialMLLM significantly outperforms baselines across both qualitative and quantitative subtasks, demonstrating an average 36\% gain and surpassing even larger proprietary models.}
\label{tab:multispa_benchmark}
\vspace{-0.7em}
\scalebox{1}{\tablestyle{3.9pt}{1}

\begin{tabular}{rcccc
>{\columncolor[HTML]{EFEFEF}}c 
>{\columncolor[HTML]{EFEFEF}}c 
>{\columncolor[HTML]{EFEFEF}}c}
\toprule
 & \textbf{Multi-SpatialMLLM} & InternVL-8B  & InternVL-26B & SpatialRGPT~\cite{SpatialRGPT} & Claude-3.5 & Gemini-2.0 & GPT-4o \\ \midrule
Average & \textbf{56.11} \improve{35.68} & 20.43 & 21.36 & 16.24 & 27.50 & 30.31 & 28.87 \\
\multicolumn{1}{l}{\cellcolor[HTML]{ECF4FF}\textit{Depth Perception}} & \multicolumn{7}{l}{\cellcolor[HTML]{ECF4FF}\textbf{}} \\
Comparison & \textbf{74.00} \improve{24.50} & 49.50 & 50.50 & 52.17 & 38.17 & 57.00 & 54.84 \\
Value & \textbf{75.33} \improve{71.99} & 3.34 & 2.50 & 3.00 & 34.84 & 28.67 & 22.50 \\
\multicolumn{1}{l}{\cellcolor[HTML]{ECF4FF}\textit{Visual Correspondence}} & \multicolumn{7}{l}{\cellcolor[HTML]{ECF4FF}\textbf{}} \\
Coordinate & \textbf{49.00} \improve{47.33} & 1.67 & 1.67 & 0.00 & 1.33 & 5.67 & 2.00 \\
MCQ & \textbf{90.00} \improve{56.67} & 33.33 & 44.00 & 30.67 & 54.67 & 73.00 & 67.67 \\
\multicolumn{1}{l}{\cellcolor[HTML]{ECF4FF}\textit{Camera Orientation}} & \multicolumn{7}{l}{\cellcolor[HTML]{ECF4FF}\textbf{}} \\
Direction & \textbf{90.83} \improve{42.66} & 48.17 & 49.34 & 54.67 & 62.17 & 62.17 & 58.84 \\
Degree & \textbf{45.50} \improve{42.16} & 3.34 & 5.17 & 4.84 & 10.50 & 16.34 & 17.50 \\
\multicolumn{1}{l}{\cellcolor[HTML]{ECF4FF}\textit{Camera Translation}} & \multicolumn{7}{l}{\cellcolor[HTML]{ECF4FF}\textbf{}} \\
Direction & \textbf{85.89} \improve{33.56} & 52.33 & 50.22 & 53.22 & 55.11 & 51.89 & 54.78 \\
Distance & \textbf{42.33} \improve{28.00} & 14.33 & 13.00 & 7.33 & 16.33 & 14.00 & 13.67 \\
Vector & \textbf{18.00} \improve{17.67} & 0.33 & 0.67 & 0.00 & 0.33 & 0.33 & 0.00 \\
\multicolumn{1}{l}{\cellcolor[HTML]{ECF4FF}\textit{Object Movement}} & \multicolumn{7}{l}{\cellcolor[HTML]{ECF4FF}\textbf{}} \\
Distance & \textbf{40.42} \improve{31.58} & 8.84 & 8.75 & 2.57 & 8.50 & 9.42 & 8.92 \\
Vector & \textbf{12.92} \improve{10.42} & 2.50 & 3.58 & 0.00 & 1.92 & 2.33 & 5.25 \\
\multicolumn{1}{l}{\cellcolor[HTML]{ECF4FF}\textit{Object Perception}} & \multicolumn{7}{l}{\cellcolor[HTML]{ECF4FF}\textbf{}} \\
Size & \textbf{49.11} \improve{21.66} & 27.45 & 26.89 & 22.33 & 46.11 & 42.89 & 40.44 \\ \bottomrule
\end{tabular}

}
\vspace{-3mm}
\end{table*}

\subsection{Multi-Frame Spatial Understanding}
\label{sec:multi-frame_spatial_understanding}

\noindent\textbf{Implementation details.}
Our preliminary studies show that InternVL2~\cite{InternLM} exhibits stronger instruction-following capabilities than other popular MLLMs (\eg, LLaVA-OneVision~\cite{LLaVA_OneVision}, VILA~\cite{VILA}). Hence, we adopt the 8B InternVL2 model as our base, fine-tuning it on the MultiSPA training split. Specifically, we employ LoRA~\cite{LoRA} with rank $R=16$ to update the LLM backbone, while freezing the image encoder and projection layer. More results of other settings are in the supplementary material. We use a cosine learning rate scheduler with $\text{lr}=4\times 10^{-5}$ and the AdamW~\cite{AdamW} optimizer. For research efficiency, we train on a subset of MultiSPA (3M QA samples) for one epoch, mixed with 60K general image-based instruction-following samples to preserve the base model's original abilities. The training is conducted on 24 nodes of 8$\times$32G V100 GPUs with a batch size of 192, taking 50 hours to complete. 

\noindent\textbf{Baselines.}~We include different sizes of InternVL2~\cite{InternLM} and SpatialRGPT~\cite{SpatialRGPT} as baselines to investigate how our proposed training data improves performance. We also evaluate three popular proprietary models, including ``Claude-3.5-Sonnet-20241022"~\cite{Claude-3.5-Sonnet}, ``Gemini2.0-Flash"~\cite{Gemini2.0-Flash}, and ``GPT-4o-20241120"~\cite{GPT-4o}, as representative models to highlight the limitations of multi-frame spatial understanding even in SOTA MLLMs. Since these baselines often either refuse to answer questions or produce responses failing to adhere to the required format, we employ prompts to encourage them to provide answers or guess values when uncertain. We further use GPT-4 for post-processing to ensure their outputs conform to the prescribed answer format and can be extracted for evaluation accordingly.

\begin{figure}[t]
\centering
\includegraphics[width=0.9\linewidth]{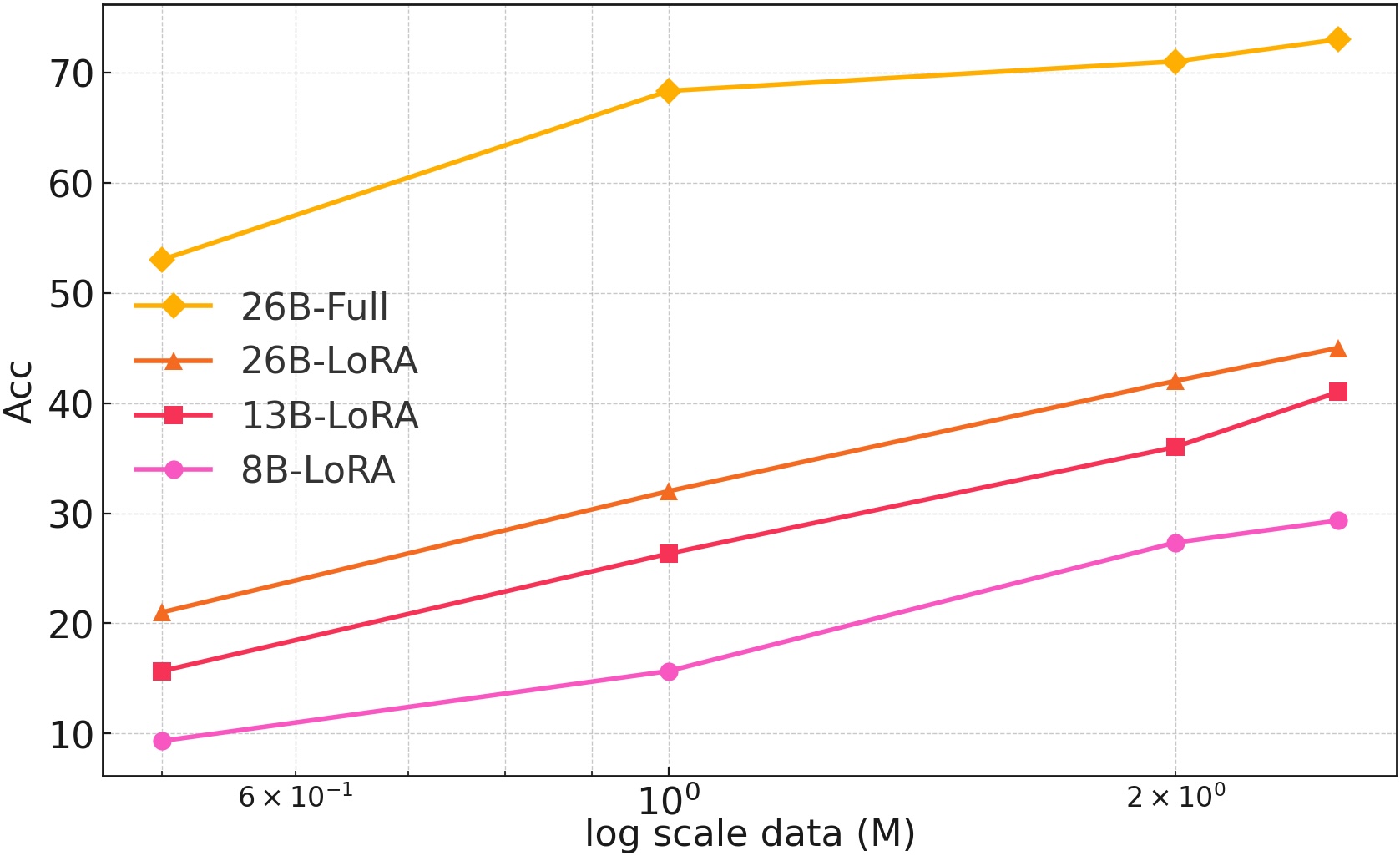}
\vspace{-2mm}
\caption{\textbf{Scalability of Multi-SpatialMLLM.}}
\vspace{-4mm}
\label{fig:scalability}
\end{figure}
% height should be 0.611 * 

\noindent\textbf{MultiSPA benchmark.}
\cref{tab:multispa_benchmark} summarizes model accuracy on our MultiSPA benchmark. We observe that most existing MLLMs have limited multi-frame spatial understanding ability, performing slightly above random (about 50–60\% accuracy) on qualitative tasks such as depth comparison and camera translation direction. 
Even worse, these baselines fail on tasks requiring quantitative outputs, such as coordinate-based visual correspondence and camera or object movement vectors.
SpatialRGPT~\cite{SpatialRGPT} is trained on vast single-image spatial data, but it performs worse than zero-shot InternVL~\cite{InternVL2_0} on MultiSPA, confirming that single-image supervision doesn't transfer to multi-image tasks.

By contrast, our Multi-SpatialMLLM significantly improves performance across all tasks, achieving an average 36\% gain over the base model. On relatively easier qualitative tasks, it gets 80–90\% accuracy and outperforms all proprietary models. Even on challenging tasks like predicting camera movement vectors, our model attains much higher accuracy, whereas all baselines remain near zero. It is notable that our model has only 8B parameters, which is likely far fewer than those of closed-source models. Yet, with the MultiSPA dataset, it matches or even exceeds their performance, validating the effectiveness of our proposed data.

\noindent\textbf{Scalability of Multi-SpatialMLLM.}
Certain tasks, like estimating the camera's displacement vector, are more difficult to learn, so using the same amount of training data brings less improvement. However, the multi-frame spatial understanding ability of Multi-SpatialMLLM is still scalable even for these challenging tasks.
To verify this, we select the camera movement vector prediction task as a case study and gradually increase the training data and trainable parameters as shown in \cref{fig:scalability}. We observe consistent improvements by adding more data and increasing model capacity. With 2.5M samples, the 26B variant achieves around 72\% accuracy. These findings encouragingly suggest that further scaling up training data and model capacity holds promise for even more powerful spatial understanding.

\noindent\textbf{Comparisons with expert models.} It's worth noting that Multi-SpatialMLLM is a general VLM, but its 3D perception capabilities can match those of SOTA specialized models. For example, our best-performing 26B model achieves 72\% accuracy on camera vector prediction, on par with VGGT~\cite{VGGT}. Additional comparisons with expert models are provided in the supplementary material.

\subsection{Generalization of Multi-SpatialMLLM}
We study the generalization ability of Multi-SpatialMLLM by evaluating it on the held-out external benchmarks and on standard VQA benchmarks. We also demonstrate the multi-task benefits introduced by our MultiSPA data.

\noindent\textbf{Held-out benchmarks.}
To verify whether our model’s multi-frame spatial understanding generalizes to other datasets, we perform zero-shot evaluation on BLINK~\cite{BLINK}, a diverse benchmark for assessing MLLM perception (\cref{tab:held_out_bench}). We focus on four splits relevant to spatial reasoning: Visual Correspondence (V.C.), Relative Depth (R.D.), Multi-View Reasoning (M.V.), and Spatial Reasoning (S.R.). Our model never sees BLINK images during fine-tuning, and BLINK's image resolutions and distributions differ from our training data. We find that all baselines fail on the M.V. task and struggle with V.C. In contrast, our Multi-SpatialMLLM achieves almost 90\% accuracy on these tasks and delivers an average 26.4\% improvement over the base model, even outperforming proprietary models. This result demonstrates that the multi-frame spatial understanding learned by our model is transferable across datasets. We do not observe gains on the S.R. task, possibly because this task focuses on topological position relations between two objects within a single image, which differs significantly from our multi-frame training data geared toward integrating spatial cues from multiple viewpoints.

To further test generalization, we also render synthetic multi-view images from unseen 3RScan~\cite{3RScan} scene meshes and evaluate our 8B and 26B models on the most challenging camera vector prediction task. Our models still achieve 18.7\% and 27.7\% accuracy, respectively, while zero-shot baselines remain at 0\%, confirming the generalization ability of Multi-SpatialMLLM.

\begin{table}[t!]
\centering
\caption{\textbf{Evaluation on the BLINK\cite{BLINK} benchmark.}}
\label{tab:held_out_bench}
\vspace{-0.7em}
\scalebox{1}{\tablestyle{4.9pt}{1.2} % {column} {row}

\begin{tabular}{lccccc}
\toprule
Model & Avg. & V.C. & R.D. & M.V. & S.R. \\ \midrule
\rowcolor[HTML]{EFEFEF} 
Gemini-2.0 & 75.7 & 88.4 & 83.9 & 42.9 & 83.9 \\
\rowcolor[HTML]{EFEFEF} 
Claude-3.5 & 67.7 & 74.4 & 63.7 & 54.1 & 75.5 \\
\rowcolor[HTML]{EFEFEF} 
GPT-4o & 73.8 & 84.9 & 71.0 & 53.4 & 81.8 \\
InternVL-8B & 57.9 & 39.0 & 71.8 & 49.6 & 76.2 \\
InternVL-26B & 61.4 & 47.1 & 78.2 & 44.4 & \textbf{79.7} \\
SpatialRGPT~\cite{SpatialRGPT} & 52.5 & 34.3 & 61.3 & 44.4 & 74.1 \\
\textbf{Multi-SpatialMLLM} & \textbf{84.3} & \textbf{89.5} & \textbf{79.8} & \textbf{94.7} & 74.8 \\ \bottomrule
\end{tabular}
}
\vspace{-1mm}
\end{table}

\begin{figure*}[t!]
\centering
\includegraphics[width=0.98\linewidth]{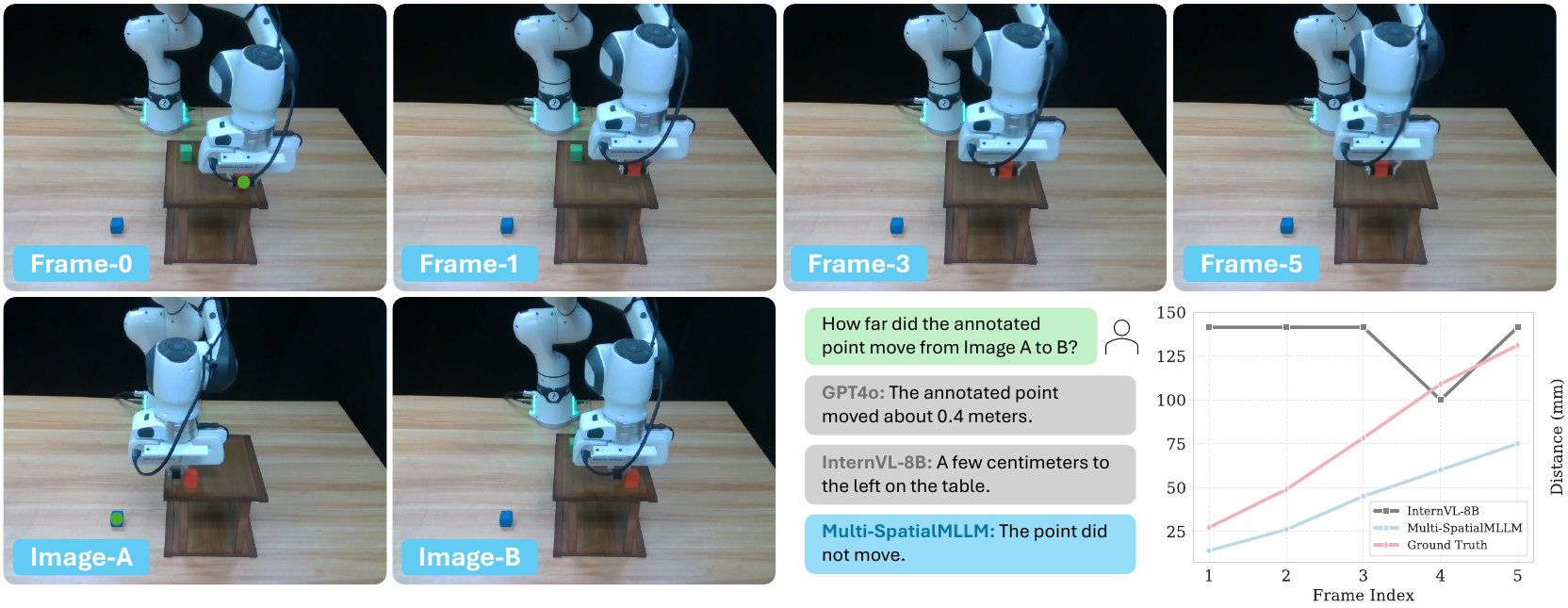}
\vspace{-3mm}
\caption{\textbf{Demonstrations of Multi-SpatialMLLM in zero-shot robotics tasks.} Our model accurately identifies static objects and predicts movement distances, aligning with the ground truth. It exhibits potential for novel applications like multi-frame reward annotation.}
\label{fig:demo}
\vspace{-3mm}
\end{figure*}

\noindent\textbf{Standard VQA benchmarks.}
We evaluate our Multi-SpatialMLLM on several popular standard VQA benchmarks, as shown in \cref{tab:general_VQA}. These benchmarks target various MLLM capabilities, such as general perception (POPE~\cite{POPE} and VizWiz~\cite{VizWiz}), optical character recognition (OCRVQA~\cite{OCRVQA}), reasoning (MathVista~\cite{MathVista} and MMStar~\cite{MMStar}), and Chinese VQA (CCBench~\cite{CCBench}). The results show rough parity across the benchmarks, indicating that our model retains most of its original standard VQA proficiency and can be used as a general-purpose MLLM, without being overfit to just multi-frame spatial reasoning.

\begin{table}[t!]
\centering
\caption{\textbf{Evaluation on standard VQA benchmarks.}}
\label{tab:general_VQA}
\vspace{-0.7em}
\scalebox{0.82}{\tablestyle{3.6pt}{0.8}

\begin{tabular}{lcccccc}
\toprule
Model & POPE & VizWiz & OCRVQA & MathVista & MMStar & CCBench \\ \midrule
InternVL-8B & 84.5 & 33.2 & 42.7 & 58.5 & 61.1 & 77.3 \\
Ours & 85.3 & 30.7 & 42.7 & 57.6 & 59.7 & 75.7 \\ \bottomrule
\end{tabular}
}
\vspace{-5mm}
\end{table}

\vspace{-1mm}
\begin{table}[h!]
\centering
\caption{\textbf{Model performance w./wo. multi-task training.}}
\label{tab:multitask}
\vspace{-0.7em}
\scalebox{0.95}{\tablestyle{4pt}{1.0}
\begin{tabular}{lcc}
\toprule
 & Camera Movement Vector & Object Movement \\ \midrule
Single Task & 9.30 & 17.50 \\
\textbf{Multiple Tasks} & \textbf{18.00 \improve{8.70}} & \textbf{22.04} \improve{4.56} \\ \bottomrule
\end{tabular}
}
\vspace{-2mm}
\end{table}

\noindent\textbf{Multi-task generalization and synergy.}
While each of the tasks proposed in \cref{sec:multispa_overview} is focused on a narrower sub-goal, ultimately the aim is to collectively improve multi-frame spatial understanding; we thus prefer that our training data has synergistic generalization effects, as opposed to balancing potentially antagonistic tasks individually. 
We observe that this is indeed the case by comparing training on just the 500K samples from the camera-movement subset (without any other task data) versus the full training set of 3M samples: we observe that the additional data from the additional tasks indeed increases the accuracy on camera movement questions from 9.3\% to 18.0\%.
We further compare two training configurations for object movement: (1) a dataset of 400K object movement samples alone, and (2) the same 400K object movement samples plus 400K additional samples from camera movement, visual correspondence, and depth estimation. The average accuracy on object movement subtasks increases from 17.5\% to 22.04\% with the additional data, as shown in \cref{tab:multitask}. Importantly, these extra 400K samples only involve ScanNet~\cite{ScanNet} images, whereas the object movement data originate from PStudio~\cite{PStudio} and ADT~\cite{ADT}, and the two sets do not share question types or data sources. This improvement demonstrates that spatial understanding learned from different datasets and task types can transfer, highlighting an additional scalability dimension beyond merely data volume and model capacity: task diversity. We provide additional results, including further verification of multi-task synergy and subset contributions, in the supplementary material.

\subsection{Emergence of spatial understanding}
We have shown that our model’s multi-frame spatial understanding is scalable (\cref{fig:scalability} and \cref{tab:multitask}). However, we also investigate whether certain spatial reasoning abilities only appear in sufficiently large models, mirroring the emergent phenomena observed in text-based LLMs~\cite{EmergentAbilitiesofLLM}.

We explore this through our multiple-choice visual correspondence task for preliminary study. By default, when generating distractor pixels in the second image, we pick them randomly; we denote this as \textbf{Easy}, as distractors may be quite far from the answer. For a more challenging scenario, we deliberately select distractors near the correct pixel, thus requiring higher discriminative power from the model (\textbf{Hard} version). We train various sizes of the base models on these Hard samples and then test on the Easy samples, to gauge whether they can effectively learn from the Hard data. \cref{tab:emergence} shows that only the 26B variant improves over the base model, whereas both the 8B and 13B models (the latter equipped with a larger 6B vision encoder) fail to learn effectively from the Hard samples and even show reduced performance. As a reference, training the 8B model on the same number of Easy samples yields 93.33\% accuracy on the test set.
These findings suggest that learning difficult spatial tasks may require sufficiently large model capacity—potentially pointing to an ``emergent" aspect of multi-frame spatial understanding. We leave deeper investigation of this interesting phenomenon to future work.

\begin{table}[t!]
\centering
\caption{\textbf{Multiple-choice visual correspondence accuracy.}}
\label{tab:emergence}
\vspace{-0.7em}
\scalebox{0.95}{\tablestyle{4pt}{1.0}

\begin{tabular}{lccc}
\toprule
Model Size & Encoder Size & LLM Size & Acc. (Baseline v.s. Hard) \\ \midrule
8B & 300M & 7B & 33.3 v.s. 25.67 \\
13B & 6B & 7B & 44.0 v.s. 42.67 \\
\textbf{26B} & \textbf{6B} & \textbf{20B} & \textbf{44.0 v.s. 82.33} \\ \bottomrule
\end{tabular}
}
\vspace{-4mm}
\end{table}

\subsection{Demonstrations of Multi-SpatialMLLM}
In \cref{fig:teaser}, we demonstrate our Multi-SpatialMLLM’s multi-view spatial understanding. We further test its real-world performance on newly collected images of a robot arm stacking cubes. These robot scenes are out-of-distribution because our training set does not include any robotic scenario. As shown in \cref{fig:demo}, when asked about the movement of a static blue cube, GPT-4o and the base model respond incorrectly, while ours accurately identifies no movement.

\noindent\textbf{Multi-Frame reward annotator.}
Prior works~\cite{SpatialRGPT, SpatialVLM} have shown that MLLMs with spatial understanding can act as reward annotators in robot learning, but they only handle single-frame inputs. In contrast, our model supports multi-frame tasks such as perceiving object movement across consecutive frames. In \cref{fig:demo}, we provide Frame-0 and subsequent frames (Frame-1 to Frame-5), then query our model about the object's displacement. Our model successfully estimates an increasing trend in movement distances, aligning with the ground truth. Though the predicted values are not exact (due to differing resolutions and domains), these results underscore our model's generalization ability and highlight potential novel applications as a reward annotator or evaluator for robot tasks involving multi-frame spatial understanding, such as ``move the object by $n$ meters."
\section{Conclusion}
\label{sec:con}
In this work, we extend MLLMs' spatial understanding to multiple frames, a capability overlooked in previous research. We develop a data generation pipeline that produces the first large-scale dataset and a benchmark, MultiSPA, dedicated to this goal. Our extensive experiments demonstrate the effectiveness, scalability, and generalization of the proposed Multi-SpatialMLLM, revealing key observations such as multi-task benefits and emergent behaviors in challenging spatial tasks. The model also opens up new applications, including acting as a multi-frame reward annotator. We discuss limitations and provide more experimental results in the supplementary material.

\clearpage
\newpage

\vspace{0.9em}
\noindent\textbf{\large{Appendix}}

\appendix
\startcontents

{
    \hypersetup{linkcolor=black}
    \printcontents{}{1}{}
}

\section{MultiSPA Data Samples and Distributions}
\label{supp_sec:data_sample}
\begin{figure*}[t!]
\centering
\includegraphics[width=0.86\linewidth]{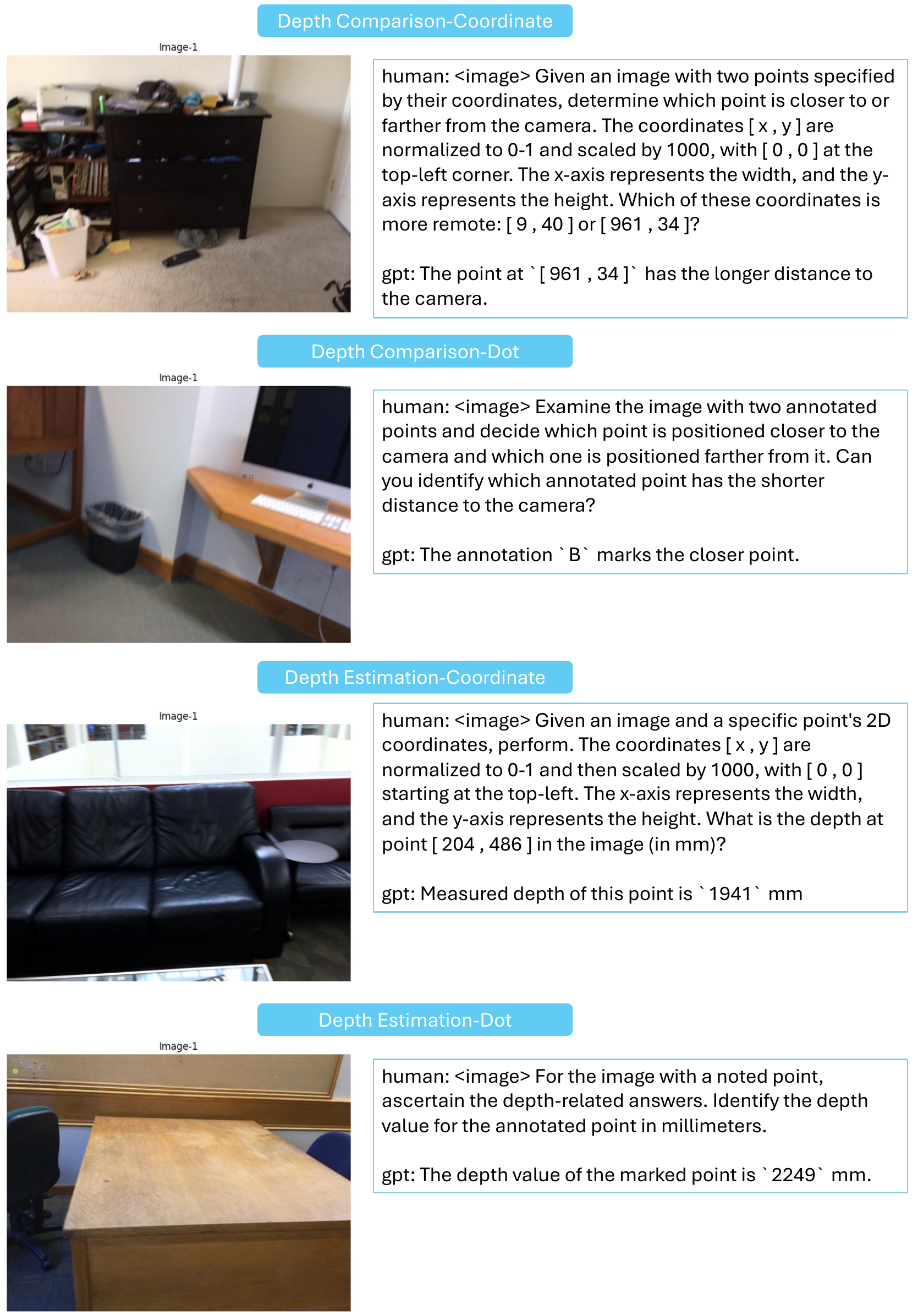}

\caption{\textbf{Data samples of depth perception.}}
\label{supp_fig:depth perception}

\end{figure*}

\begin{figure*}[t!]
\centering
\includegraphics[width=0.86\linewidth]{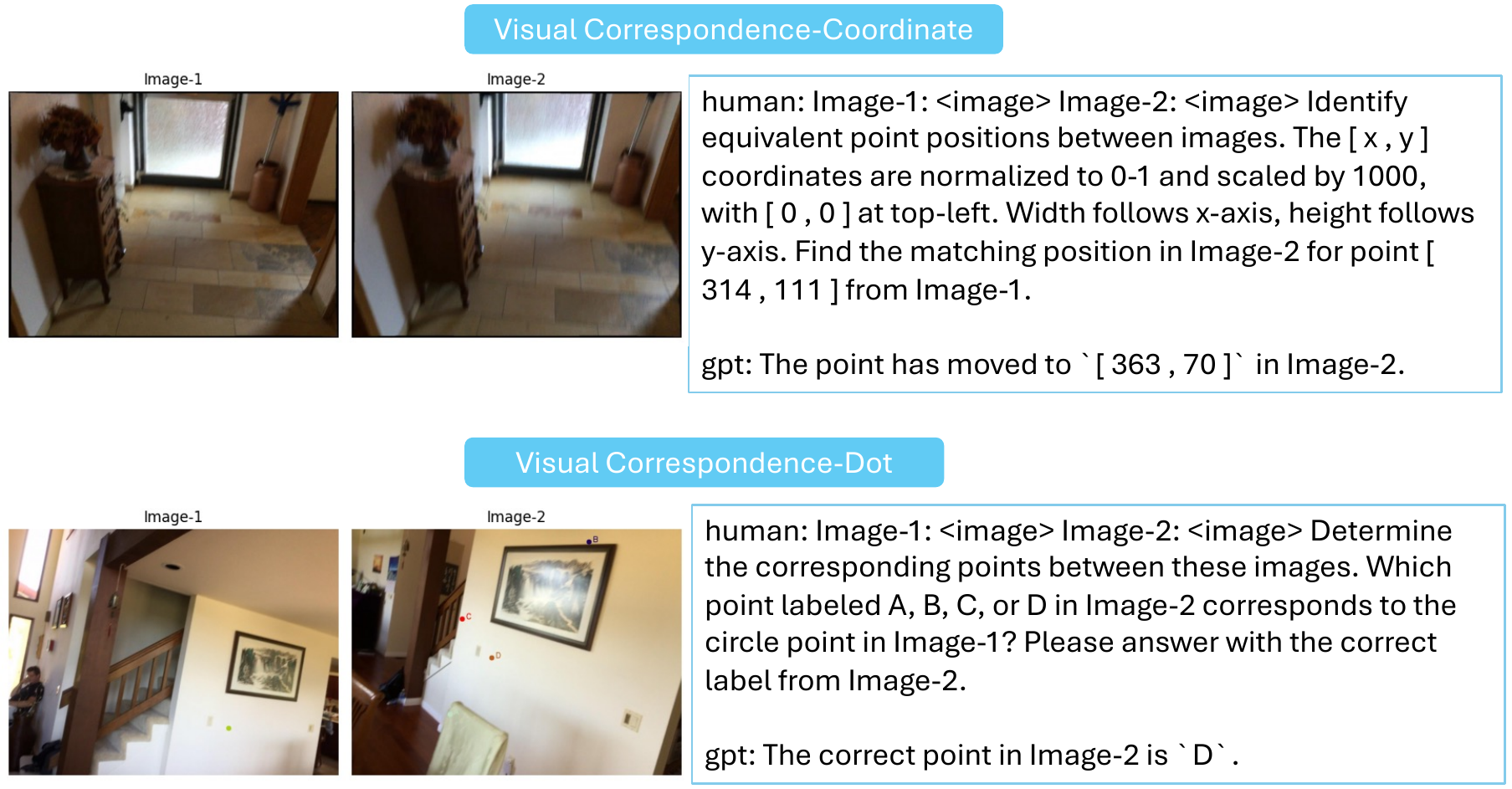}

\caption{\textbf{Data samples of visual correspondence.}}
\label{supp_fig:visual correspondence}

\end{figure*}

\begin{figure*}[t!]
\centering
\includegraphics[width=0.86\linewidth]{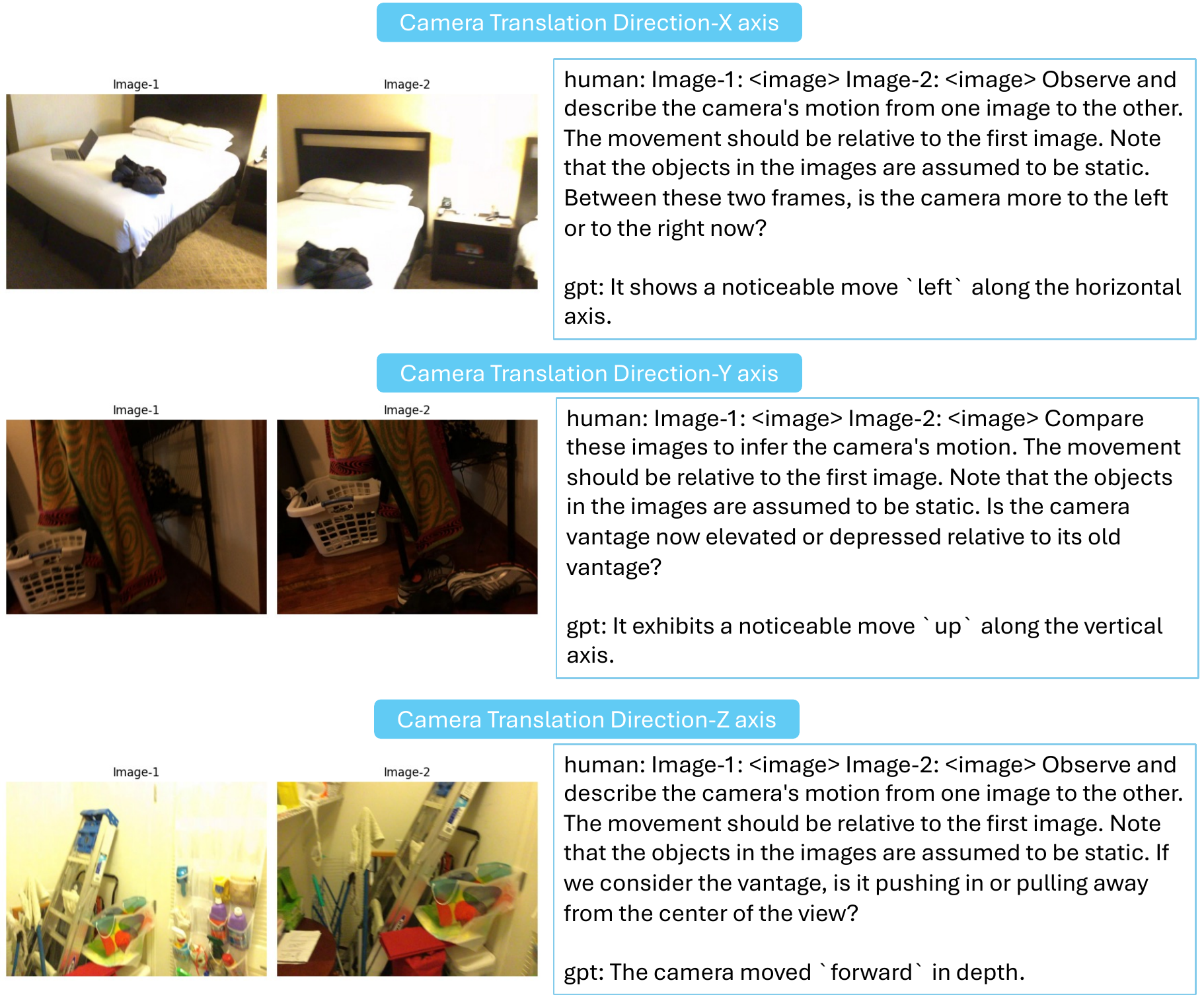}

\caption{\textbf{Data samples of camera movement-translation direction.}}
\label{supp_fig:camera movement-translation direction}

\end{figure*}

\begin{figure*}[t!]
\centering
\includegraphics[width=0.86\linewidth]{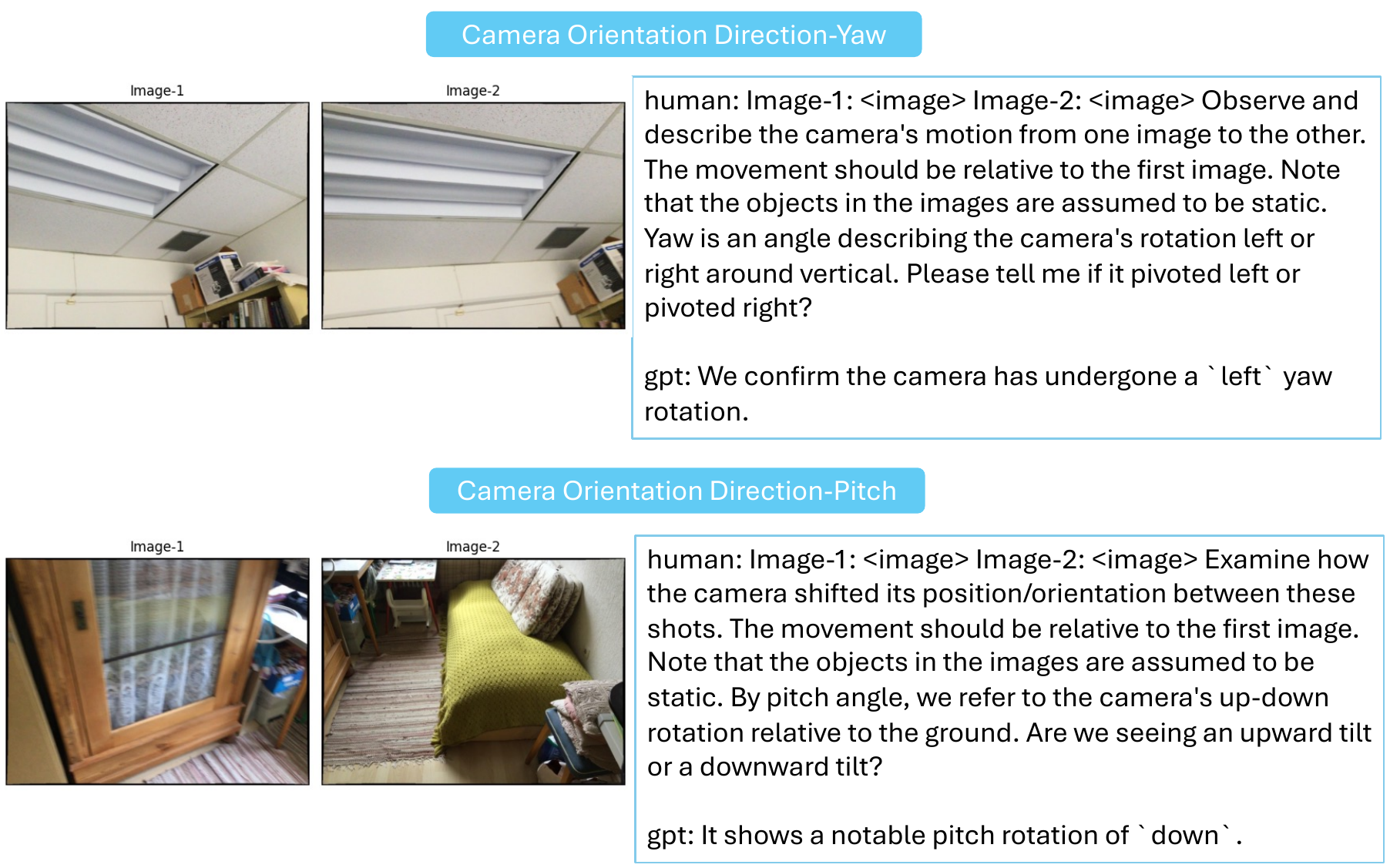}

\caption{\textbf{Data samples of camera movement-orientation direction.}}
\label{supp_fig:camera movement-orientation direction}

\end{figure*}

\begin{figure*}[t!]
\centering
\includegraphics[width=0.86\linewidth]{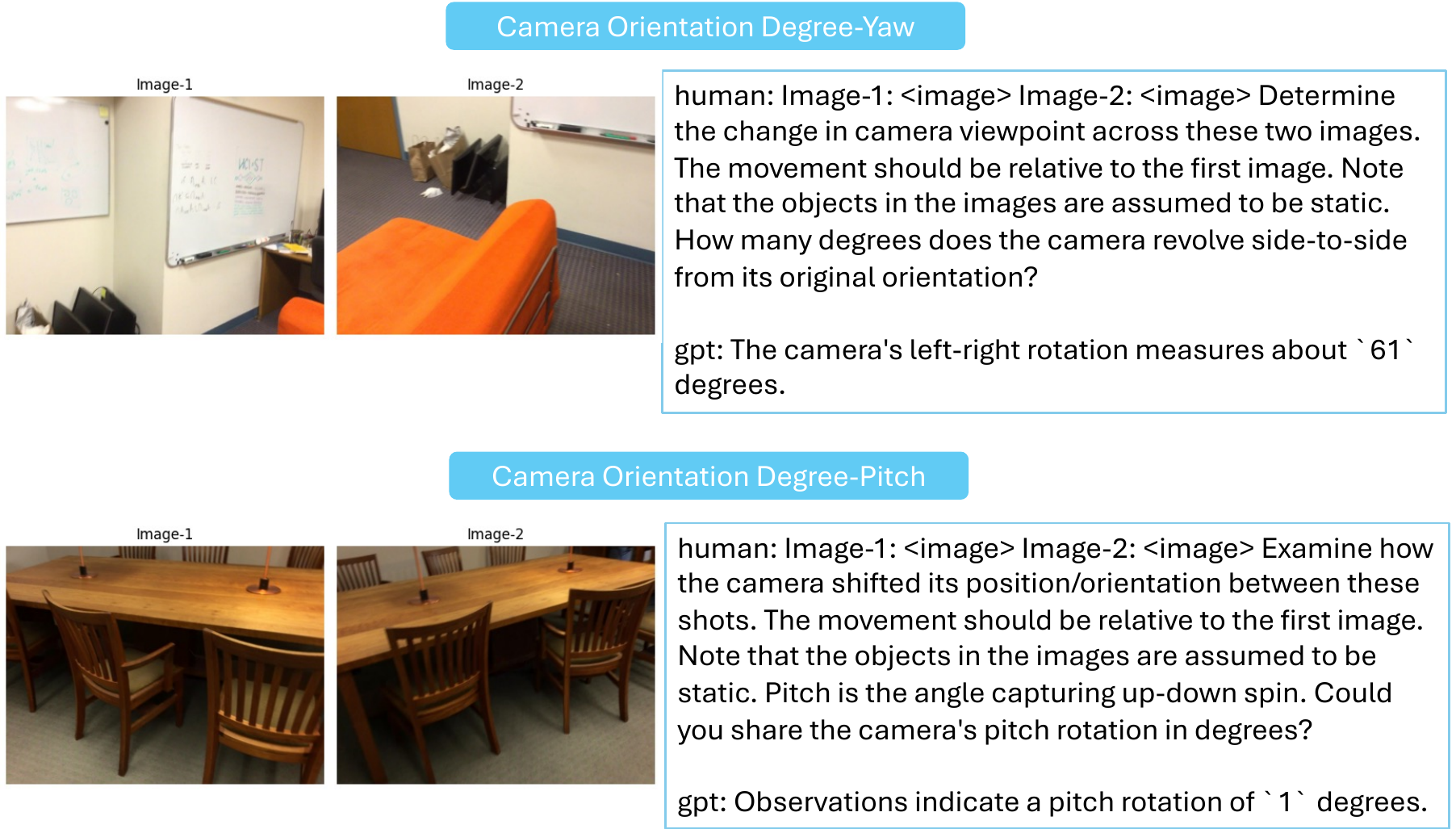}

\caption{\textbf{Data samples of camera movement-orientation degree.}}
\label{supp_fig:camera movement-orientation degree}

\end{figure*}

\begin{figure*}[t!]
\centering
\includegraphics[width=0.86\linewidth]{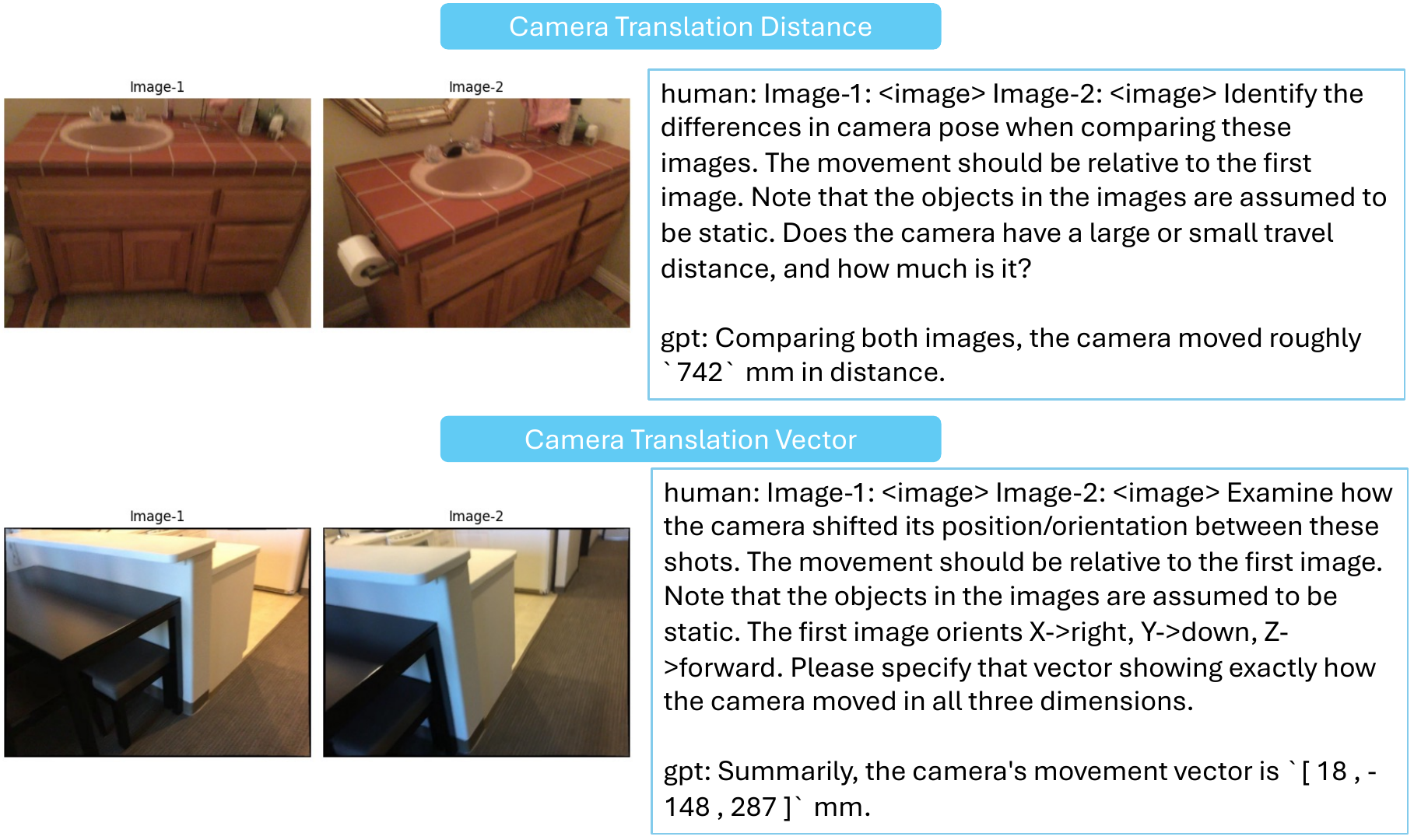}

\caption{\textbf{Data samples of camera movement-translation distance and vector.}}
\label{supp_fig:camera_translation_distance_vector}

\end{figure*}

\begin{figure*}[t!]
\centering
\includegraphics[width=0.86\linewidth]{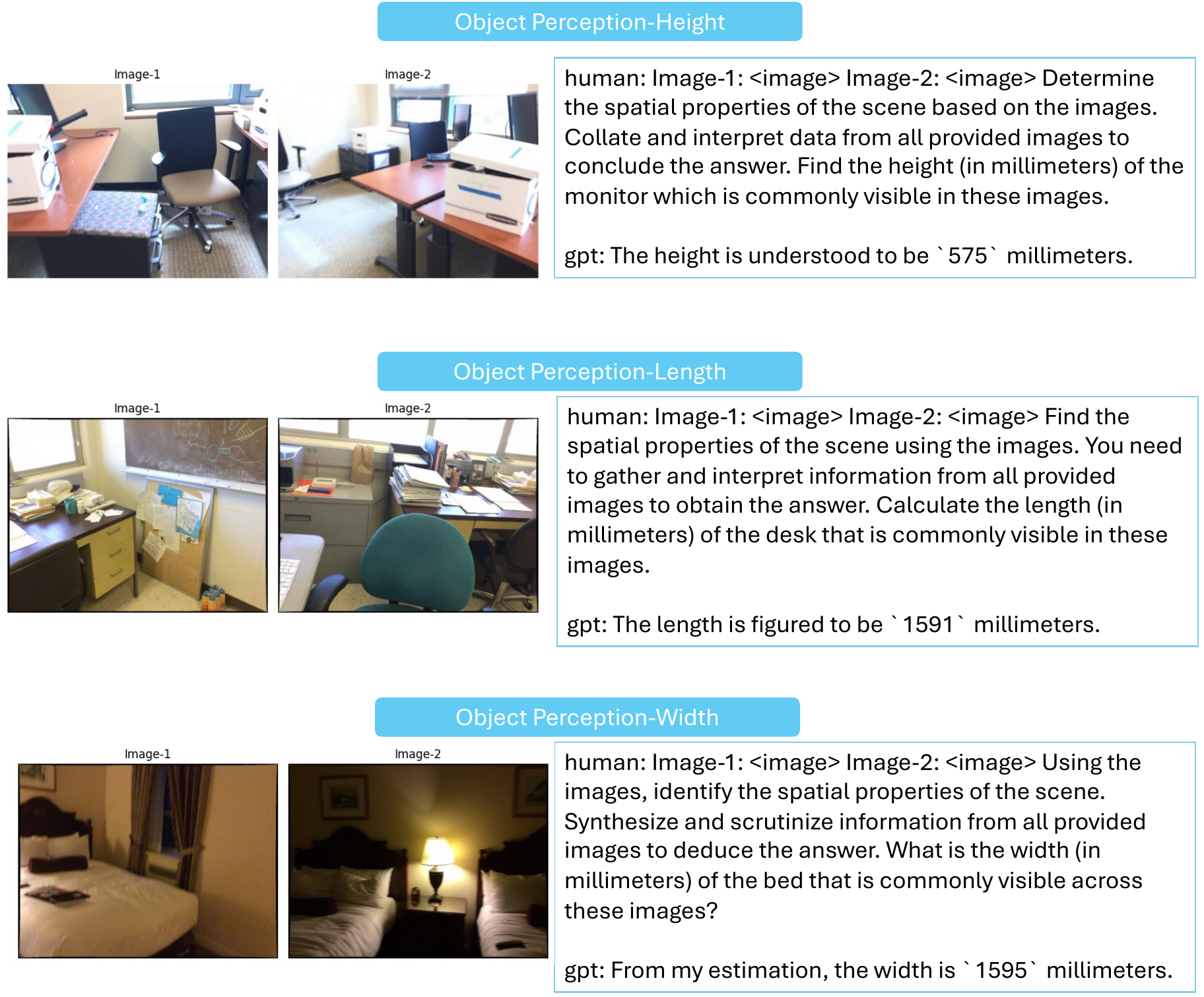}

\caption{\textbf{Data samples of object perception.}}
\label{supp_fig:object_perception}

\end{figure*}

\begin{figure*}[t!]
\centering
\includegraphics[width=0.86\linewidth]{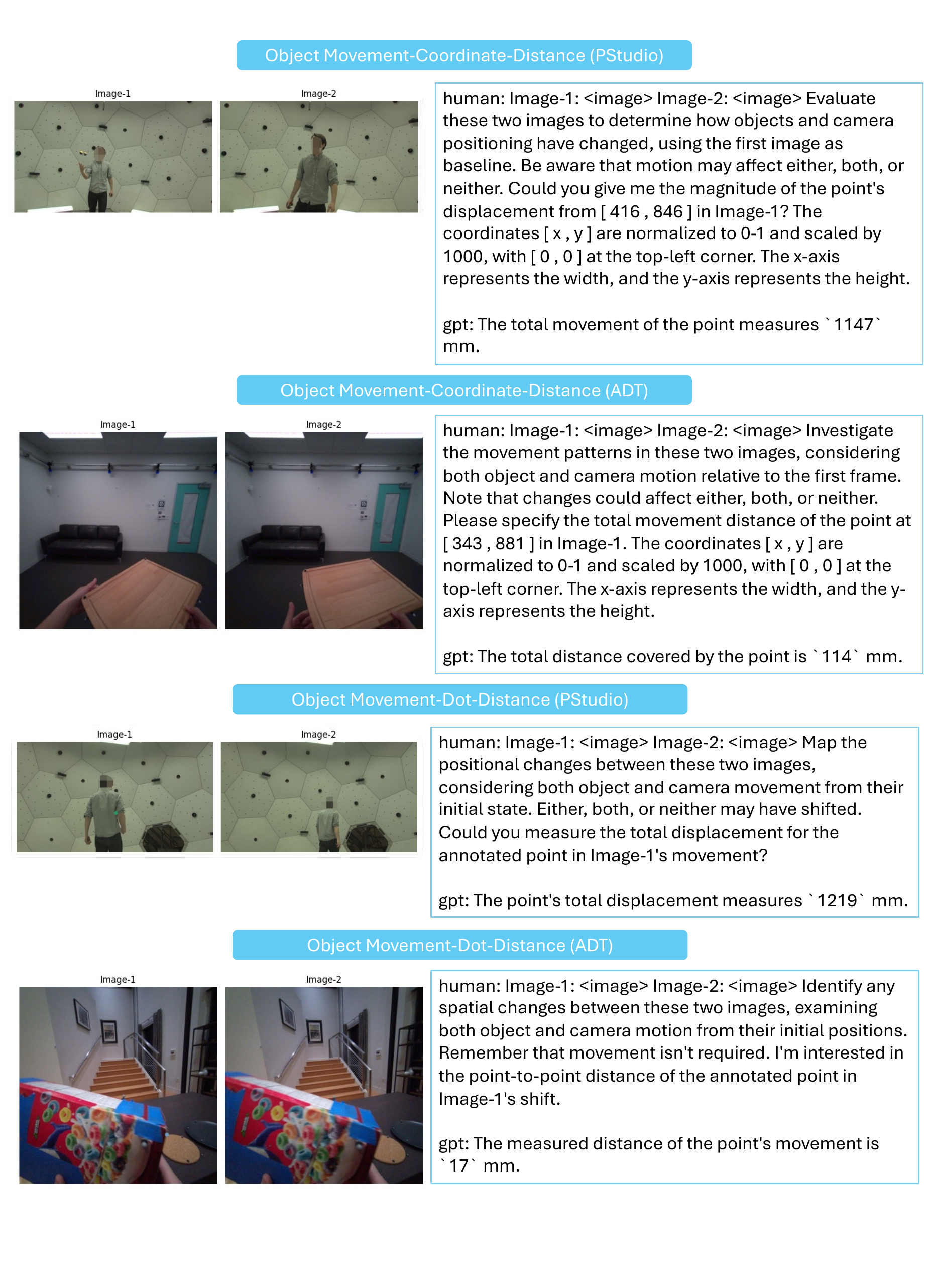}

\caption{\textbf{Data samples of object movement-distance.}}
\label{supp_fig:object_movement_distance}

\end{figure*}

\begin{figure*}[t!]
\centering
\includegraphics[width=0.86\linewidth]{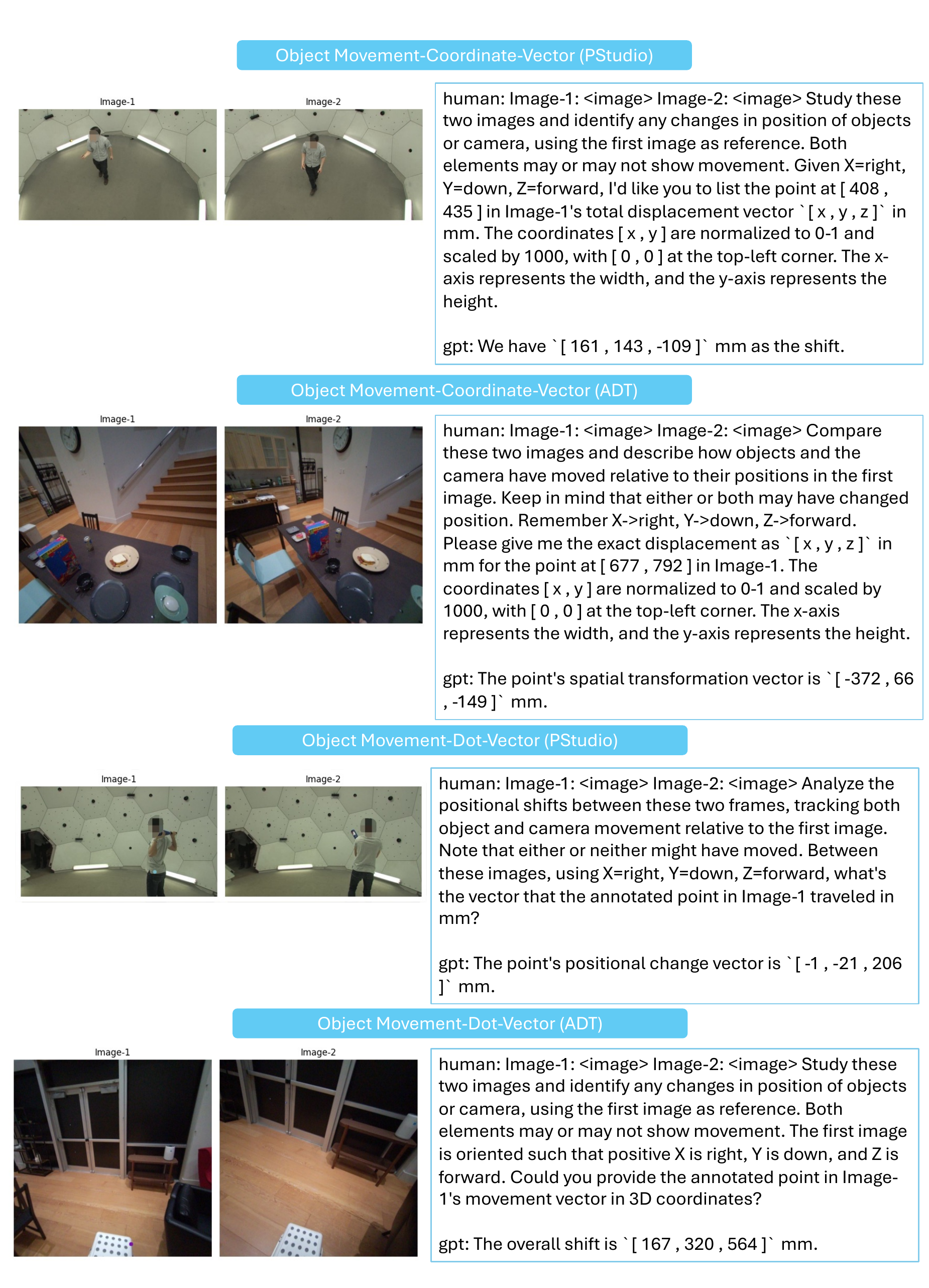}

\caption{\textbf{Data samples of object movement-vector.}}
\label{supp_fig:object_movement_vector}

\end{figure*}

Our MultiSPA dataset has 26 subtasks in total. Each task with an example is shown from \cref{supp_fig:depth perception} to \cref{supp_fig:object_movement_vector}.
We show the distribution of samples across different task types in \cref{supp_tab:data_dist}. Note that our data engine is scalable, and we can generate more training samples for these tasks.

\begin{table*}[t]
\centering
\caption{\textbf{Distribution of samples across different task types.}}
\label{supp_tab:data_dist}
\vspace{-0.7em}
\scalebox{0.98}{\tablestyle{3.6pt}{0.8}

\begin{tabular}{cccccc}
\toprule
Depth Estimation & Visual Correspondence & Camera Orientation & Camera Translation & Object Movement & Object Perception \\ \midrule
1.6M & 1.5M & 4M & 9M & 13M & 1.5M \\ \bottomrule
\end{tabular}
}
% \vspace{-5mm}
\end{table*}

\begin{table*}[th]
\centering
\caption{\textbf{Results of different fine-tuning strategies for InternVL2-8B.} Increasing the number of trainable parameters yields larger gains from MultiSPA. Tuning the vision encoder significantly improves performance on fine-grained spatial understanding tasks, such as predicting camera motion vectors.}

\label{supp_tab:different_fine-tuning}
\vspace{-0.7em}
\scalebox{1}{\tablestyle{3.9pt}{1}

\begin{tabular}{r
c
c 
c 
c}
\toprule
 & LLM-LoRA & LLM-LoRA+MLP  & LLM-LoRA+MLP+VE & LLM-Full+MLP+VE \\ \midrule
Average & 56.11 & 57.46 & 65.41 & \textbf{66.72} \\
\multicolumn{1}{l}{\cellcolor[HTML]{ECF4FF}\textit{Depth Estimation}} & \multicolumn{4}{l}{\cellcolor[HTML]{ECF4FF}\textbf{}} \\
Comparison & 74.00 & 73.34 & 81.00 & 80.84 \\
Value & 75.33 & 75.84 & 78.50 & \textbf{79.67} \\
\multicolumn{1}{l}{\cellcolor[HTML]{ECF4FF}\textit{Visual Correspondence}} & \multicolumn{4}{l}{\cellcolor[HTML]{ECF4FF}\textbf{}} \\
Coordinate & 49.00 & 52.67 & 63.67 & \textbf{71.33} \\
MCQ & 90.00 & 90.33 & 94.67 & \textbf{95.67} \\
\multicolumn{1}{l}{\cellcolor[HTML]{ECF4FF}\textit{Camera Orientation}} & \multicolumn{4}{l}{\cellcolor[HTML]{ECF4FF}\textbf{}} \\
Direction & 90.83 & 91.50 & 94.00 & \textbf{94.17} \\
Degree & 45.50 & 46.84 & 59.67 & \textbf{60.50} \\
\multicolumn{1}{l}{\cellcolor[HTML]{ECF4FF}\textit{Camera Translation}} & \multicolumn{4}{l}{\cellcolor[HTML]{ECF4FF}\textbf{}} \\
Direction & 85.89 & 87.66 & 91.78 & \textbf{92.56} \\
Distance & 42.33 & 46.33 & 72.00 & \textbf{77.67} \\
Vector & 18.00 & 24.67 & 43.00 & \textbf{53.00} \\
\multicolumn{1}{l}{\cellcolor[HTML]{ECF4FF}\textit{Object Movement}} & \multicolumn{4}{l}{\cellcolor[HTML]{ECF4FF}\textbf{}} \\
Distance & 40.42 & 38.17 & 39.08 & \textbf{47.83} \\
Vector & 12.92 & 15.84 & 19.50 & \textbf{26.75} \\
\multicolumn{1}{l}{\cellcolor[HTML]{ECF4FF}\textit{Object Perception}} & \multicolumn{4}{l}{\cellcolor[HTML]{ECF4FF}\textbf{}} \\
Size & 49.11 & 46.33 & 48.00 & 47.22 \\ \bottomrule
\end{tabular}

}
\vspace{-3mm}
\end{table*}

\section{MultiSPA Data Templates}
Due to paper length limits, we only show part of the templates in \cref{code:templates}. Other templates are similar to those shown in this supplementary material. 

\section{Details of Source Datasets}

\noindent\textbf{ScanNet.}
ScanNet \cite{ScanNet} is an RGB-D dataset containing more than 1,500 indoor scans. Each scan provides reconstructed point clouds, 3D camera poses, camera intrinsics, depth maps, and 3D instance and semantic segmentation masks. Our data generation pipeline utilizes all these annotations, though segmentation masks are optional if object perception data is not required.

\noindent\textbf{PStudio.}
The CMU Panoptic Studio dataset \cite{PStudio} comprises 65 sequences (5.5 hours total) of multiple people interacting with one another or with objects, captured within a light stage. It offers multi-view images, 3D body skeletons, and facial landmark annotations.

\noindent\textbf{ADT.}
Aria Digital Twin \cite{ADT} is an egocentric video dataset recorded with Aria glasses. It contains 200 sequences of real-world indoor activities, each with precise 6DoF camera poses, 3D human poses, 2D image segmentations, and depth maps, as well as a digital twin environment.

\noindent\textbf{TAPVid3D.}
TAPVid3D~\cite{TAPVid3D} is a dataset for tracking 3D points in space. It provides temporal 3D point tracking, constructed from PStudio, ADT, and DriveTrack \cite{DriveTrack}. It leverages official annotations to produce temporally aligned 3D point sequences, along with camera pose sequences and intrinsics. We use these annotations for our data generation. Note that we exclude the DriveTrack split because its camera poses are insufficiently accurate.

\section{Image Pairs Sampling}

To ensure a balanced selection of image pairs based on their overlap ratio, we adopt the procedure as follows. First, we separate pairs with zero overlap and randomly sample a predefined number of them. Next, we partition all nonzero-overlap pairs into bins according to their overlap ratio. We then distribute the sampling quota across bins in proportion to the number of bins, sorting them by bin size in ascending order to prevent smaller bins from being overshadowed by larger ones. Finally, we either sample or exhaust each bin, carrying over any unused quota to subsequent bins. This step balances pairs of different overlap levels, mitigating issues caused by long-tail distributions. The main content of the full algorithm is shown in \cref{code:overlap_sampling}.

\section{Rotation Angles}
Beyond translation, we estimate two orientation angles: yaw and pitch. We do not model roll, as it typically remains small in real-world use cases (\eg, autonomous vehicles, robotics, wearable devices). Formally, let \(\mathbf{E}\in\mathbb{R}^{4\times4}\) be the camera pose in world coordinates, which has the $z-$axis aligned with the gravity direction, and \(\mathbf{R}\) its upper-left \(3\times3\) submatrix:
\begin{equation}
\mathbf{R} =
\mathbf{E}[0{:}3,0{:}3].
\label{eq:rotation_submatrix}
\end{equation}
We then extract yaw and pitch by focusing on the camera’s forward (i.e., \(z\)-) axis:
\begin{equation}
\mathbf{z}_{\text{fwd}} = \mathbf{R} \begin{bmatrix}0 \\ 0 \\ 1\end{bmatrix}.
\label{eq:rotated_z_axis}
\end{equation}
\noindent\textbf{Yaw} is defined as the angle of this rotated \(z\)-axis in the horizontal plane, measured around the gravity axis:
\begin{equation}
\text{yaw} = 
\arctan2\bigl(\mathbf{z}_{\text{fwd}}[1],\,\mathbf{z}_{\text{fwd}}[0]\bigr)\times\frac{180}{\pi}.
\label{eq:yaw_angle}
\end{equation}
\noindent\textbf{Pitch} is the angle of \(\mathbf{z}_{\text{fwd}}\) relative to the ground plane:
\begin{equation}
\text{pitch} = 
\arcsin\!\Bigl(\frac{\mathbf{z}_{\text{fwd}}[2]}{\|\mathbf{z}_{\text{fwd}}\|}\Bigr)\times\frac{180}{\pi}.
\label{eq:pitch_angle}
\end{equation}
With these two angles, we can determine whether the camera rotates left or right, and tilt up or tilt down.

\section{BFS-Based Minimum-Coverage-Set Search}
\label{sec:coverage_search}
To ensure that an object’s full dimensions are captured across multiple images, we develop a breadth-first search (BFS) algorithm that identifies minimal sets of images whose combined coverage meets each dimension’s size requirement. In particular, for each axis (height, width, length), we track which subsets of object points are visible per image. If the difference between the minimum and maximum coordinates of all selected points along that axis meets a target threshold (based on the object’s 3D bounding box size), we consider it covered. Our BFS proceeds in two phases at each iteration: 
\begin{enumerate}
    \item \emph{Phase A: Coverage check.} We examine the current sets and mark any that fully cover the object on the chosen axis. These sets are recorded as ``minimal,'' and any set that is a superset of a previously found minimal set is pruned. 
    \item \emph{Phase B: Expansion.} We expand the remaining (uncovered) sets to the next level by appending additional images, while pruning those that cannot possibly achieve coverage in deeper levels.
\end{enumerate}
This process continues until either no further expansion is possible or the maximum number of images is reached. The final result is a collection of minimal sets that together span the object’s relevant dimension. Although we include pruning steps, the search still becomes expensive when considering sets of three or more images. Hence, we only use two images for object size perception in our data. \Cref{code:coverage_search} shows a simplified implementation.

\section{Clustering-Based Rigid Body Segmentation}
\label{sec:rigid_body_seg}
In TAPVid3D~\cite{TAPVid3D}, all points in a sequence often belong to the same object or scene region, but they can be unevenly distributed (\eg, a human torso versus arms). To sample diverse motion patterns, we segment the point cloud into multiple rigid bodies, each undergoing a distinct motion. Our method accumulates inter-point distance changes over time and applies hierarchical clustering to identify coherent groups. We also filter groups with too less points to avoid noise. \Cref{code:rigid_seg} is a simplified code snippet.

\section{Different Fine-Tuning Strategies}
For research efficiency, we adopt a resource-light setting: we fine-tune the LLM backbone with LoRA while freezing the vision encoder and the MLP projection layer. This limits the trainable components and thus the benefits from our proposed datasets. We also evaluate alternative fine-tuning strategies; results are shown in \cref{supp_tab:different_fine-tuning}.

We observe that increasing the number of trainable parameters yields larger gains from MultiSPA. Notably, tuning the vision encoder significantly improves performance, especially on challenging fine-grained spatial perception tasks such as camera and object motion prediction. For example, enabling vision-encoder tuning increases accuracy on camera translation vector prediction from 24.67\% to 43.00\%, highlighting the importance of the vision encoder for learning spatially relevant visual representations. These results are consistent with the scalability trends reported in the main paper.

\section{Comparison with Expert Models}
Multi-SpatialMLLM is a general-purpose VLM capable of broad vision–language tasks. Enabled by MultiSPA data and training, it acquires multi-frame spatial understanding, such as camera motion estimation, depth perception, and visual correspondence (image matching), traditionally achieved by specialized 3D vision models. We compare these capabilities to expert systems; results are summarized in \cref{supp_tab:comparison_with_expert}.

For camera vector prediction, we compare to the SOTA VGGT~\cite{VGGT} using our fully fine-tuned 26B variant trained on 2.5M camera-vector samples. Because VGGT does not output metric-scale vectors, we report results after scale normalization. For the depth comparison task, we evaluate against Depth-Anything~\cite{DepthAnything}; for image matching, we evaluate against LoFTR~\cite{LoFTR}, using our fully fine-tuned 8B variant trained on 3M mixed MultiSPA samples. We could not include a 26B fully fine-tuned model on the mixed MultiSPA set due to computational limits at submission.

Nevertheless, our models match VGGT and LoFTR, and our 8B model remains competitive with Depth-Anything. Based on the scaling trends reported in the main paper, we expect further gains with larger models and data. Our goal is not to surpass specialized systems, but to show that a general-purpose VLM can attain comparable spatial perception performance while retaining broad vision–language capabilities.

\vspace{-1mm}
\begin{table}[h!]
\centering
\caption{\textbf{Comparison with expert models.} For the camera vector prediction task, we compare VGGT to our 26B model; for depth comparison and image matching, we compare to our 8B model.}
\label{supp_tab:comparison_with_expert}
\vspace{-0.7em}
\scalebox{0.95}{\tablestyle{4pt}{1.0}
\begin{tabular}{ccc}
\toprule
 VGGT v.s. Ours & Depth Anything v.s. Ours & LoFTR v.s. Ours \\ \midrule
86 v.s. 82 & 86 v.s. 76 & 71 v.s. 71 \\ \bottomrule
\end{tabular}
}
\vspace{-2mm}
\end{table}

\section{More on VQA Benchmarks}
\noindent\textbf{More results on spatial benchmarks.} We extend our evaluation to held-out spatial benchmarks, including the popular CVBench-3D~\cite{CVBench} and ERQA~\cite{ERQA}. As shown in \cref{supp_tab:more_spatial_VQA}, our model outperforms the baseline on both benchmarks, with particularly strong gains on the ERQA multi-frame subset, confirming its effective generalization to unseen benchmarks.

\begin{table}[ht]
\centering
\caption{\textbf{Evaluation on spatial benchmarks.}}
\label{supp_tab:more_spatial_VQA}
\vspace{-0.7em}
\scalebox{0.85}{\tablestyle{3.6pt}{0.7}

\begin{tabular}{lccc}
\toprule
Model & CVBench-3D~\cite{CVBench} & ERQA~\cite{ERQA} & ERQA: M.V.~\cite{ERQA} \\ \midrule
InternVL-8B & 78.2 & 35.8 & 13.5 \\
Multi-SpatialMLLM & 81.7 & 36.2 & 21.6 \\ \bottomrule
\end{tabular}
}
% \vspace{-5mm}
\end{table}

\noindent\textbf{More results on general benchmarks.} In addition to evaluating our models on the general VQA benchmarks reported in the main paper, we further test them on three more widely-used VQA benchmarks to verify that our fine-tuning preserves most of the general VLM capabilities, as shown in \cref{supp_tab:more_general_VQA}.

\begin{table}[ht]
\centering
\caption{\textbf{Evaluation on standard VQA benchmarks.}}
\label{supp_tab:more_general_VQA}
\vspace{-0.7em}
\scalebox{0.9}{\tablestyle{3.6pt}{0.8}

\begin{tabular}{lccc}
\toprule
Model & MMMU~\cite{MMMU} & MME~\cite{MME} & MMVet~\cite{MMVet} \\ \midrule
InternVL-8B & 47.7 & 85.8 & 62.0 \\
Multi-SpatialMLLM & 48.4 & 84.9 & 58.1 \\ \bottomrule
\end{tabular}
}
% \vspace{-5mm}
\end{table}

\begin{figure}[t]
\centering
\includegraphics[width=0.9\linewidth]{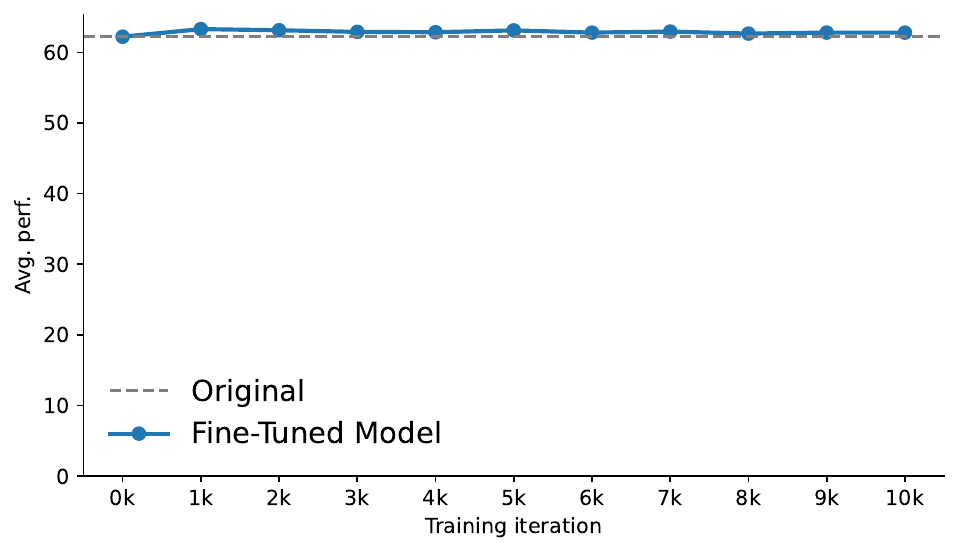}
\vspace{-2mm}
\caption{\textbf{Forgetting curves of Multi-SpatialMLLM training.}}
% \vspace{-4mm}
\label{supp_fig:forgetting_curve}
\end{figure}

\noindent\textbf{Forgetting curve.} We also track the average performance on all 10 general VQA benchmarks throughout training, as shown in \cref{supp_fig:forgetting_curve}. The results demonstrate that our model maintains stable performance, alleviating concerns about forgetting. We hypothesize that the learned spatial skills are largely orthogonal to general capabilities, thereby minimizing interference with the model’s pre-existing knowledge during fine-tuning.

\section{Multi-Task Synergy}
\noindent\textbf{Further verification.} To further validate the benefits of multi-task training, we use the multiple-choice visual correspondence (V.C.) task, which provides a strong baseline (random guess: 25\%). We report results under different cross-task training configurations in \cref{supp_tab:mutlitask}.

We find that including a small set of target-task samples, 1K V.C. examples, which alone are insufficient to improve accuracy, enables the model to leverage knowledge from other tasks (50K samples each). Experiments show that adding data from camera movement, object perception, and depth perception consistently boosts V.C. performance, further confirming the multi-task synergy highlighted in the main paper.

\vspace{-1mm}
\begin{table}[h!]
\centering
\caption{\textbf{Multi-task synergy for visual correspondence (V.C.).} Incorporating training samples from other tasks improves V.C. performance. Abbreviations: Cam., camera movement; Obj., object perception; Depth., depth perception.}
\label{supp_tab:mutlitask}
\vspace{-0.7em}
\scalebox{0.95}{\tablestyle{4pt}{1.0}
\begin{tabular}{ccccc}
\toprule
 Zero-Shot & V.C. & V.C. + Cam. & V.C. + Obj. & V.C. + Depth. \\ \midrule
33.3 & 33.1 & 40.6 & 41.0 & 57.3 \\ \bottomrule
\end{tabular}
}
\vspace{-2mm}
\end{table}

\noindent\textbf{Benefits of different tasks.} We also observe that depth perception contributes the largest gains to V.C. The main paper shows that adding samples from other datasets helps, but it also remains unclear which task contributes most.

To study this, we adopt the same multi-task setting but include 400K training samples from a single task at a time, V.C., depth perception, or object perception. As shown in \cref{supp_tab:subset_contribution}, V.C. and depth samples yield the largest improvements, reinforcing the benefits of multi-task training and suggesting that V.C. and depth perception may be more fundamental spatial tasks, consistent with the classic Structure-from-Motion pipeline~\cite{MVG}.

\vspace{-1mm}
\begin{table}[h!]
\centering
\caption{\textbf{Benefits of different tasks.} “Single” denotes training without samples from other tasks. Abbreviations: V.C., visual correspondence; Depth., depth perception; Obj., object perception.}
\label{supp_tab:subset_contribution}
\vspace{-0.7em}
\scalebox{0.95}{\tablestyle{4pt}{1.0}
\begin{tabular}{ccccc}
\toprule
& Single & w. V.C. & w. Depth. & w. Obj. \\ \midrule
 Camera Movement & 9.30 & 15.00 & 15.00 & 12.00 \\
Object Movement & 17.50 & 21.88 & 22.59 & 18.54 \\ \bottomrule
\end{tabular}
}
\vspace{-2mm}
\end{table}

\section{Generalization to More Images.}
In the main paper, we primarily train on two-frame samples. To test generalization to more frames, we evaluate camera direction prediction by asking fully fine-tuned models to predict motion between the first and last frame while inserting 1–4 intermediate frames. As shown in \Cref{supp_tab:varying_frames}, accuracy remains at 85\%, demonstrating robustness and generalization to varying frame counts.

\vspace{-1mm}
\begin{table}[h!]
\centering
\caption{\textbf{Performance of camera prediction when adding different numbers of frames.}}
\label{supp_tab:varying_frames}
\vspace{-0.7em}
\scalebox{1}{\tablestyle{4pt}{1.0}
\begin{tabular}{ccccc}
\toprule
 \# Frames & 1 & 2 & 3 & 4 \\ \midrule
Acc. & 85.0 & 85.3 & 85.3 & 85.3 \\ \bottomrule
\end{tabular}
}
\vspace{-2mm}
\end{table}

\section{Limitations}
\label{sec:limitations}
Our generated data in the main paper uses two-frame scenarios. We show that models trained on two images generalize to more images, and our data-generation pipeline naturally scales to additional frames. However, future work is still needed to extend beyond pairs to exploit multi-view inputs for stronger spatial reasoning. Another limitation is that although we observe signs of the emergent phenomenon, further investigation is required to clarify what exact spatial abilities drive such emergence.

\clearpage
\newpage
\newpage
\begin{figure*}
\begin{lstlisting}[caption=The image pairs sampling algorithm for static scene data, label=code:overlap_sampling]

def sample_dataframe(df, all_overlap_samples, non_overlap_samples,
                     overlap_min=0, overlap_max=100, interval=1):
    # 1) Sample pairs with overlap == 0
    non_overlap_df = df[df["overlap"] == 0].copy()
    sampled_non_overlap_df = (non_overlap_df if len(non_overlap_df) <= non_overlap_samples
                              else non_overlap_df.sample(n=non_overlap_samples))

    # 2) Partition the remaining pairs (overlap != 0) into bins
    remaining_df = df[df["overlap"] != 0].copy()
    bins = np.arange(overlap_min, overlap_max + interval, interval)
    remaining_df["overlap_group"] = pd.cut(remaining_df["overlap"], bins=bins, include_lowest=True)
    remaining_df.dropna(subset=["overlap_group"], inplace=True)

    bin_groups = []
    for ovlp_bin, group_df in remaining_df.groupby("overlap_group"):
        bin_groups.append((ovlp_bin, group_df))
    if not bin_groups:
        final_df = sampled_non_overlap_df.copy()
        final_df.drop(columns=["overlap_group"], errors="ignore", inplace=True)
        return final_df

    # 3) Distribute all_overlap_samples evenly across bins
    N = len(bin_groups)
    base_quota = all_overlap_samples // N
    remainder = all_overlap_samples % N
    bin_quotas = [base_quota] * N
    for i in range(remainder):
        bin_quotas[i] += 1

    # 4) Sort bins by size (ascending) and sample
    bin_data = []
    for i, (ovlp_bin, group_df) in enumerate(bin_groups):
        bin_data.append({
            "group_df": group_df,
            "quota": bin_quotas[i],
            "size": len(group_df)
        })
    bin_data.sort(key=lambda x: x["size"])

    sampled_df = pd.DataFrame()
    leftover = 0
    for info in bin_data:
        group, quota, size = info["group_df"], info["quota"], info["size"]
        current = quota + leftover
        if size <= current:
            sampled_df = pd.concat([sampled_df, group], ignore_index=True)
            leftover = current - size
        else:
            sampled_df = pd.concat([sampled_df, group.sample(n=current)], ignore_index=True)
            leftover = 0

    if leftover > 0:
        print(f"Warning: leftover {leftover} samples not used.")

    # 5) Combine sampled bins with zero-overlap samples
    final_df = pd.concat([sampled_df, sampled_non_overlap_df], ignore_index=True)
    final_df.drop(columns=["overlap_group"], errors="ignore", inplace=True)
    return final_df
\end{lstlisting}
\end{figure*}
\newpage
\begin{figure*}
\begin{lstlisting}[language=Python, caption={Simplifed version of BFS-based minimum-coverage-set search with pruning.}, label=code:coverage_search]
def compute_coverage(points, mask, axis):
    """Returns the min-to-max spread along 'axis' for points indicated by 'mask'."""
    if not mask.any():
        return 0.0
    coords = points[mask][:, axis]
    return coords.max() - coords.min()

def covers_dimension(coverage, target_dim, tol):
    """Checks if 'coverage' is within tolerance of the target dimension."""
    return abs(coverage - target_dim) <= tol * target_dim

def bfs_min_coverage(images, visibility, points, obj_mask, axis, target_dim, tol, max_k=2):
    """
    Finds minimal image sets up to size 'max_k' that meet coverage criteria along 'axis'.
    'images' is a list of candidate frames, 'visibility' maps frame->boolean mask,
    'obj_mask' indicates the object points in 'points'.
    """
    # Prepare BFS queue: each item is (set_of_images, combined_mask, last_idx)
    queue = []
    for i, img in enumerate(images):
        mask_i = visibility[img] & obj_mask
        queue.append(([img], mask_i, i))

    solutions = []
    k = 1
    while k <= max_k and queue:
        next_level = []
        for combo, comb_mask, last_idx in queue:
            cov = compute_coverage(points, comb_mask, axis)
            if covers_dimension(cov, target_dim, tol):
                solutions.append(combo)
            elif k < max_k:
                # Expand only if we have not reached max_k
                for j in range(last_idx + 1, len(images)):
                    mask_j = visibility[images[j]] & obj_mask
                    next_mask = comb_mask | mask_j
                    next_level.append((combo + [images[j]], next_mask, j))
        queue = next_level
        k += 1
    return solutions
\end{lstlisting}
\end{figure*}
\newpage
\begin{figure*}
\begin{lstlisting}[language=Python, caption={Rigid body segmentation with smoothing and hierarchical clustering.}, label=code:rigid_seg]
def smooth_distance_changes(dist_t, dist_prev, smooth_factor=0.01):
    """Zeroes out small distance changes to reduce noise."""
    diff = np.abs(dist_t - dist_prev)
    return np.where(diff > smooth_factor, diff, 0)

def rigid_body_segmentation(points, thr=0.1, smooth_factor=0.01):
    """
    points: Shape (T, N, 3), with T time steps & N points.
    thr: Threshold for clustering distance.
    smooth_factor: Ignored small changes.
    Returns: A list of groups, each group is a list of point indices.
    """
    T, N, _ = points.shape
    cum_loss = np.zeros((N, N))

    # Accumulate distance changes over time
    for t in range(1, T):
        dist_t = squareform(pdist(points[t]))
        dist_prev = squareform(pdist(points[t - 1]))
        cum_loss += smooth_distance_changes(dist_t, dist_prev, smooth_factor)

    # Hierarchical clustering
    Z = linkage(squareform(cum_loss), method="average")
    labels = fcluster(Z, thr, criterion="distance")

    # Group points by label
    groups = []
    for label_id in range(1, labels.max() + 1):
        group = np.where(labels == label_id)[0].tolist()
        groups.append(group)
    return groups
\end{lstlisting}
\end{figure*}
\newpage
\begin{figure*}
\begin{lstlisting}[caption=Part of the templates used by MultiSPA dataset, label=code:templates]
# Depth Estimation-Dot
TASK_DESCRIPTION = [
    "<image>\nGiven an image with an annotated point, complete the question-answer task.",
]
TEMPLATES = {
    "questions": [
        "What is the depth of the annotated point in the image (in mm)?",
    ],
    "answers": [
        "The depth of the annotated point is `{depth}` mm.",
    ]
}

# Visual Correspondence Multiple-Choice
TASK_DESCRIPTION = [
    "Image-1: <image>\nImage-2: <image>\nGiven these two images, find the corresponding points between them.",
]

TEMPLATES = {
    "questions": [
        "Which point labeled A, B, C, or D in Image-2 corresponds to the circle point in Image-1? Please answer with the correct label from Image-2.",
    ],
    "answers": [
        "The correct point is labeled `{correct_label}`.",
    ]
}

# Object Perception
TASK_DESCRIPTION = [
    "Assume the scene remains unchanged. Your task is to determine the spatial properties based on the images. You need to integrate and analyze information from all provided images to get the answer.",
]

QUESTION_TEMPLATES = [
    "What is the {dimension} (in millimeters) of the {object_category} itself commonly visible in these images?",
]

ANSWER_TEMPLATES = [
    "The {dimension} is approximately `{value_mm}` millimeters.",
]

# Object Movement-Coordinate-Distance

TASK_DESCRIPTION = [
    "Image-1: <image>\nImage-2: <image>\nGiven two images, analyze the movements of objects in the images and the cameras that captured them. The movement should be relative to the first image. Note that the objects in the images and the camera may or may not have moved.",
]

QUESTION_TEMPLATES = [
    "How far did the point at [ {x1} , {y1} ] in Image-1 travel between the two shots? The coordinates [ x , y ] are normalized to 0-1 and scaled by 1000, with [ 0 , 0 ] at the top-left corner. The x-axis represents the width, and the y-axis represents the height."
]

ANSWER_TEMPATES =  [
    "The point traveled a total of `{total_distance}` mm.",
]

\end{lstlisting}
\end{figure*}

\clearpage

{
    \small
    \bibliographystyle{ieeenat_fullname}
    \bibliography{main}
}

\end{document}